\documentclass{article}

\usepackage{PRIMEarxiv}

\usepackage[utf8]{inputenc} 
\usepackage[T1]{fontenc}    
\usepackage{hyperref}       
\usepackage{url}            
\usepackage{booktabs}       
\usepackage{amsfonts}       
\usepackage{nicefrac}       
\usepackage{microtype}      
\usepackage{fancyhdr}       
\usepackage{graphicx}       
\graphicspath{{media/}}     
\usepackage{calc}
\usepackage{amsmath}
\usepackage{amssymb}
\usepackage{geometry}
\usepackage{longtable}
\usepackage{multirow}
\usepackage{caption}
\usepackage{subcaption}

\newcommand{\nnmatwithdims}[3]{$#1 \in \mathbb{R}_{+}^{#2 x #3}$}

\newcommand*{\horzbar}{\rule[.5ex]{8.0ex}{0.5pt}}
\newcommand*{\smallvertbar}{\rule[-0.25ex]{0.5pt}{2.0ex}}
\newcommand*{\smallhorzbar}{\rule[.5ex]{1.0ex}{1.0pt}}

\DeclareMathOperator{\tfidf}{score_{tf-idf}}
\DeclareMathOperator{\tf}{TF}
\DeclareMathOperator{\idf}{IDF}

\DeclareMathOperator{\dirichlet}{Dirichlet}
\DeclareMathOperator{\multinom}{Multinomial}
\DeclareMathOperator{\rank}{rank}
\DeclareMathOperator{\similarity}{similarity}
\DeclareMathOperator{\softmax}{softmax}
\DeclareMathOperator{\attention}{\text{att}}
\DeclareMathOperator{\mattention}{\text{MultiheadedAtt}}

\DeclareMathOperator*{\argmax}{argmax}

\DeclareMathOperator{\indicatorop}{\mathbb{I}}
\newcommand{\indicator}[1]{\indicatorop\{#1\}}
\newcommand{\diffnorm}[2]{ \Vert #1 - #2 \Vert }

\newcommand{\attentionp}[1]{\attention(QW_{#1}^{Q}, KW_{#1}^{K}, VW_{#1}^{V})}
\newcommand{\dirichletp}[1]{\dirichlet(#1)}
\newcommand{\multinomp}[1]{\multinom(#1)}

\newcommand{\pparam}[2]{p\left(#1;#2\right)}

\title{No Pattern, No Recognition: a Survey about Reproducibility and Distortion Issues of Text Clustering and Topic Modeling
}

\begin{document}

\author{
  \small{Marília Costa Rosendo Silva, Felipe Alves Siqueira, João Pedro Mantovani Tarrega}\\
  \small{Institute of Mathematics and Computer Science}\\
  \small{University of São Paulo}\\
  \small{Av. Trabalhador São-Carlense, 400}, {13566-590}\\
  \small{São Carlos, São Paulo, Brazil}\\
  \texttt{\{marilia.costa.silva,felipe.siqueira,joao.tarrega\}@usp.br}\\
  \and
  \small{\textbf{João Vitor Pataca Beinotti, Augusto Sousa Nunes, Miguel de Mattos Gardini}}\\
  \small{Institute of Mathematics and Computer Science}\\
  \small{University of São Paulo}\\
  \small{Av. Trabalhador São-Carlense, 400}, {13566-590}\\
  \small{São Carlos, São Paulo, Brazil}\\
  \texttt{\{joaobeinotti,augustonunes,miguelgardini\}@usp.br}\\  
  \and
  \small{\textbf{Vinicius Adolfo Pereira da Silva}}\\
  \small{Institute of Mathematics and Computer Science and São Carlos School of Engineering}\\
  \small{University of São Paulo}\\
  \small{Av. Trabalhador São-Carlense, 400}, {13566-590}\\
  \small{São Carlos, São Paulo, Brazil}\\
  \texttt{vinicius.adolfo.silva@usp.br} \\
   \and
  \small{{\textbf{Nádia Félix Felipe da Silva}}}\\
  \small{Institute of Mathematics and Computer Science / Institute of Informatics}\\
  \small{University of São Paulo / Federal University of Goiás}\\
  \small{Av. Trabalhador São-Carlense, 400, 13566-590} / {Alameda Palmeiras, Quadra D - Campus Samambaia, 74690-900}\\
  \small{São Carlos, São Paulo, Brazil} / {Goiânia, Goiás, Brazil}\\
  \texttt{nadia.felix@ufg.br} \\
  \and
  \small{\textbf{André Carlos Ponce de Leon Ferreira de Carvalho}}\\
  \small{Institute of Mathematics and Computer Science}\\
  \small{University of São Paulo}\\
  \small{Av. Trabalhador São-Carlense, 400}, {13566-590}\\
  \small{São Carlos, São Paulo, Brazil}\\
  \texttt{andre@icmc.usp.br} \\
  }

\maketitle

\vspace{-0.8cm}

\begin{abstract}
	Extracting knowledge from unlabeled texts using machine learning algorithms can be complex. Document categorization and information retrieval are two applications that may benefit from unsupervised learning (e.g., text clustering and topic modeling), including exploratory data analysis. However, the unsupervised learning paradigm poses reproducibility issues. The initialization can lead to variability depending on the machine learning algorithm. Furthermore, the distortions can be misleading when regarding cluster geometry. Amongst the causes, the presence of outliers and anomalies can be a determining factor. Despite the relevance of initialization and outlier issues for text clustering and topic modeling, the authors did not find an in-depth analysis of them. This survey provides a systematic literature review (2011-2022) of these subareas and proposes a common terminology since similar procedures have different terms. The authors describe research opportunities, trends, and open issues. The appendices summarize the theoretical background of the text vectorization, the factorization, and the clustering algorithms that are directly or indirectly related to the reviewed works. 
\end{abstract}

\keywords{text \and natural language processing \and clustering \and topic modeling \and state-of-the-art \and survey}

\section{Introduction}
\label{sec:introduction}

Big Data, coined in the mid-1990s, involves different data sets (e.g., transactions, text, audios, videos, and images) generated in a high velocity, high volume, wide variety and can deliver data-driven decision-making means to enterprises and academia \cite{GANDOMI2015BIGDATA}. In 2021, users posted 575,000 tweets every minute \cite{Domo2021data}. Thus, deriving information from large corpora without text mining techniques (e.g., natural language processing (NLP) and machine learning) is infeasible \cite{GANDOMI2015BIGDATA}.

NLP involves processing text and speech generated by humans with artificial intelligence-based and computational resources \cite{jurafsky2021speech}. Speech recognition, automatic translation, opinion mining, and text clustering are some NLP tasks. In this work, the authors constrained the systematic literature review (SLR), based on an adaptation of Preferred Reporting Items for Systematic reviews and Meta-Analyses (PRISMA) \cite{Page2021}, to the text clustering task (which includes topic modeling).

Albeit the availability of large corpora, labeled text is scarce. The text clustering task can group instances or documents even though they are unlabeled. Document categorization \cite{Zhu2013}, information retrieval \cite[p.350]{Manning2008}, and sentiment analysis \cite{Sharuee2018} are some applications of text clustering. Topic modeling (as part of text clustering \cite[p.107]{AggarwalZhai2012}) also comprises how the groups are connected and how they change over time \cite{grootendorst2020bertopic}. In the introduction, we unified the meaning of topic modeling and text clustering. Nonetheless, in the following sections, they can be discussed individually.

The topic modeling and text clustering research areas have some imprecise terminologies. They can be represented by one of the most important concepts when involving clustering tasks: the definition of a cluster \cite{Castro2002, Hennig2015}. In a general sense, clusters are thought of as groups of similar instances or partitions of the data into more homogeneous subsets based on some criteria. Clustering algorithms assume one way to approximate human intuition regarding how the data should be grouped, varying between individuals and task contexts. Due to this subjectivity, the clustering algorithm objective function may not match our intuitions. Depending on the assumptions about data, it would not be possible to solve optimally, which can be hard to validate in high dimensions. Thus, these premises can directly affect the quality of clusters or topics \cite{harrando2021}.

Topic modeling can assign meaning to the clusters. This task involves the application of statistical algorithms to find patterns that frequently appear in a collection of documents. According to \cite{Blei2012}, topic modeling algorithms can be applied to different data types, such as images, genetic sequences, social networks, and texts. When applied to the text corpora, topic modeling algorithms look for topics, also called themes. The topics are probability distributions over words or $n$-grams\footnote{Sequence of $n$ adjacent tokens in a text.}.

The literature does not present a unified description with respect to the terminology. Technique \cite{Alghamdi2015,hoyle2021automated}, algorithm \cite{allahyari2017brief}, tool \cite{Chang2009}, probabilistic model \cite{Lafferty2006}, NLP task \cite{harrando2021}, and method \cite{JelodarLatent2018} can all be called topic modeling. Henceforth, the initialization problem \cite{langville2014}, the outlier detection \cite{Mohotti2020}, the topic modeling \cite{Blei2003}, and the text clustering \cite[p. 77-78]{AggarwalZhai2012} are referred to as tasks. Analogously, the means that perform these tasks are referred to as algorithms in this survey.
\begin{figure}
    \includegraphics[width=0.70\textwidth]{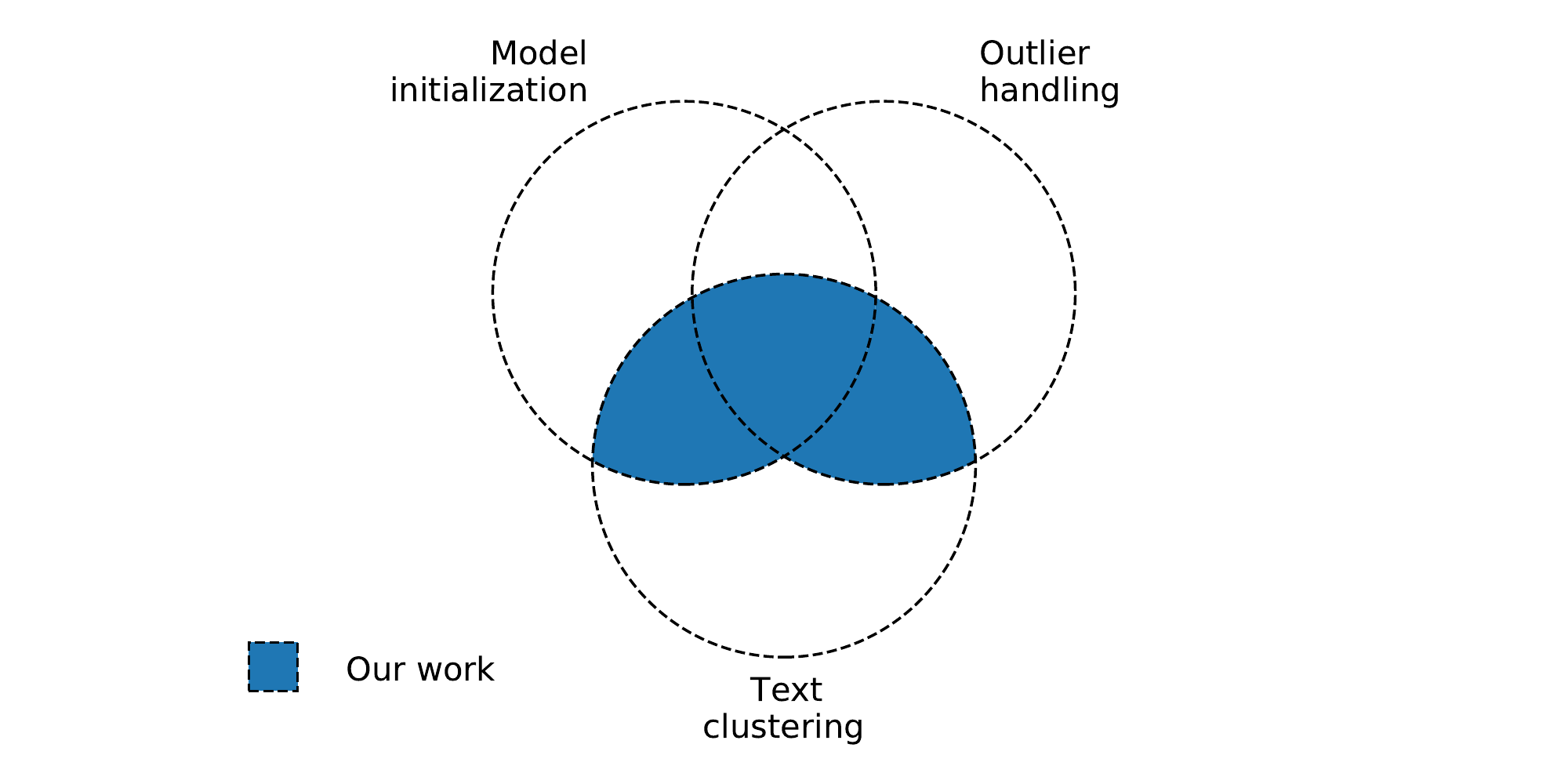}
    \centering
    \caption{Venn diagram depicting how the present work is inserted into the central themes of ``model initialization'', ``outlier handling'' and ``text clustering'', indicated by the filled regions.}
    \label{fig:work-concepts}
\end{figure}
In this survey, we address two concerns in text clustering and topic modeling: initializing algorithms and detecting and handling unusual observations, commonly known as outliers or anomalies \cite{Salgado2016}. Figure \ref{fig:work-concepts} presents the Venn diagram of the scope of this literature review.
Topic modeling handles several data types and can be part of frameworks (e.g., BERTopic \cite{grootendorst2020bertopic}). However, we considered in Figure \ref{fig:work-concepts} that topic modeling is part of text clustering.

Some algorithms are sensitive to initial conditions or outliers, and the results may fall short of our expectations, or the optimization convergence rate may be compromised with weak initialization or biased toward a small number of instances whenever there are poor outlier detection and handling. The definition of ``data outlier'' is context-dependent \cite{Chandola2009, Wang2019c, Aggarwal2013}, and ``noise'', ``anomaly'' or many other terms are often treated as synonyms, although their exact meaning is not consensual \cite{Carreo2019}. Concerning topic modeling, ``outliers'' can also be defined at different levels: token, sentence, or topic level, which requires distinct approaches. While often thought of as undesired noise in data, many outliers need to be appropriately addressed, as they also represent the corpora structure, such as underrepresented topics \cite{Veselova2020} or rapidly emerging topics \cite{Schubert2014}.

The benchmarks should provide sufficient variability (and difficulty levels) to assess the strengths and weaknesses of each outlier detection algorithm \cite{Campos2016evaluationofunsupervised, Ruff2021unifyingreview} and fully recognize breakthroughs and advancements in the literature \cite{emmot2015metaanalysis, Ruff2021unifyingreview}. Nevertheless, without robust benchmarks and with a smaller variability of datasets, the results can be excessively optimistic \cite{emmot2015metaanalysis}. The benchmarks also must describe clear and concise evaluation criteria \cite{Ruff2021unifyingreview}, and new algorithms may be tested using larger testbeds \cite{Han2022ADBench}. Notwithstanding the importance of the benchmarks, there is an absence of standardized evaluation metrics, inconsistencies in evaluation procedures, benchmarks, and pipelines \cite{zhao-ijcai2021-638,harrando2021,terragni-etal-2021-octis,lisena-etal-2020-tomodapi,hoyle2021automated,doogan-buntine-2021-topic}. 

There are efforts to define the state-of-the-art (SOTA) of topic models \cite{harrando2021}. However, establishing the SOTA can be misleading when the existing metrics cannot be suitable for the newest topic modeling algorithms \cite{doogan-buntine-2021-topic}. Due to the ambiguity of human judgment toward automated metrics (e.g., coherence scores), it is uncertain whether a positive correlation between human evaluation and automatic metrics implies causality. Nonetheless, the inconsistencies in performance evaluation metrics are out of the scope of this survey.

To the best of our knowledge, there are neither surveys nor reviews concerning outlier detection and how to improve the quality of the initialization for text data for text clustering and topic modeling \cite{AggarwalZhai2012,Alghamdi2015,allahyari2017brief,Burkhardtsurvey2019,Qiangshorttext2020,zhao-ijcai2021-638}. Thus, this systematic literature review discusses outliers and initialization issues in the aforementioned context.


\noindent Our research questions are:
\begin{enumerate}
    \item Which are the terminologies of the clustering, topic modeling, initialization, and outlier detection literature?
    \item How does the literature address the randomness and, consequently, the reproducibility loss associated with initialization issues?
    \item How does the literature address the distortion associated with outlier detection issues?
    \item Which are the key factors that compromise scientific cooperation and decelerate the advancement of the research fields?
    \item What research opportunities are there?
\end{enumerate}

\noindent Our contributions to this work are:
\begin{enumerate}
    \item To adapt the PRISMA SLR method to the domain of computer science research;
    \item To summarize found initialization algorithms and outlier detection algorithms concerning topic modeling and clustering;
    \item To clarify the terminology of the text clustering and adjacent subjects;
    \item To present and summarize text clustering algorithms with their: theoretical background, adjacent algorithms, computational complexities, and limitations;
    \item To point out research opportunities, and open issues for text clustering and topic modeling.
\end{enumerate}

This work is organized as follows: Section \ref{sec:methods} presents the information related to how we selected our reviewed works, the eligibility criteria, and synthesizes relevant characteristics from the selected works. Section \ref{sec:existing-benchmarks} details the existing benchmarks with respect to outlier detection. Section \ref{sec:results-outlier-handling} disambiguates outliers and details the algorithms that handle them. In Section \ref{sec:results-initialization}, there are presented initialization issues and algorithms to mitigate the risk of weak initialization. Section \ref{sec:end-to-end-topic-models} briefly reviews current end-to-end topic models. Section \ref{sec:discussion} presents general discussions of the reviewed works, research opportunities, and currently open issues. Section \ref{sec:related-works} describes other literature reviews for topic modeling and clustering. Appendix \ref{sec:text-vectorization} presents the methods the reviewed algorithms used to vectorize the text data before the clustering or the topic modeling. Finally, appendices \ref{sec:results-factorization-methods} and \ref{sec:results-clustering-methods} introduce the foundations of factorization and clustering algorithms related to the reviewed works, respectively.

\section{Methodology for the Work Selection}\label{sec:methods}

The SLRs enable the summarization of the current state of the literature concerning a predefined scope \cite{Page2021}. When synthesizing these studies, research opportunities and gaps in previous research can be identified, compared to primary research that does not answer these problems alone \cite{Page2021}. Despite the fact that PRISMA focuses on health sciences, this work adapted the method to computer science domain. Moreover, the PRISMA-P \cite{Shamseerg2015protocol} provides a review protocol that enhances reproducibility, defines the scope of the review, provides guidelines to write the article, reduces the partiality, and plans the decision-making (e.g., eligibility and exclusion criteria, and data extraction).

In particular, this survey could not perform meta-analyses, which consist of statistical analysis aiming to generalize the results of independent studies \cite{Shamseerg2015protocol}, because the literature of topic modeling and clustering does not use comparable metrics, and there was no consensus about the most appropriate metrics to compare all approaches \cite{zhao-ijcai2021-638,harrando2021,terragni-etal-2021-octis,lisena-etal-2020-tomodapi,hoyle2021automated,doogan-buntine-2021-topic}.

This section is divided as follows. Subsection \ref{Information-Sources} presents our information sources, whereas the following defines the eligibility criteria and search strategy. The Subsection \ref{subsec:data-collection} presents our data extraction method, and the Subsection \ref{subsec:synthesis-reviewed} summarizes information from our data extraction.

\subsection{Information Sources}\label{Information-Sources}

This literature review included three sources: ACLweb\footnote{\url{https://aclanthology.org/}}, Scopus\footnote{\url{https://www.scopus.com/home.uri}} and Web of Science\footnote{\url{https://www.webofscience.com/}}. The ACLweb (anthology of the Association for Computational Linguistics (ACL)) comprises conference papers on computational linguistics. Scopus is a search engine that includes conferences and journals. There are rigorous criteria to select and index these publishers. Most of the highest cited conferences according to the H5-Index\footnote{\url{https://scholar.google.com/citations?view_op=top_venues}} of Google Scholar are amongst the indexed proceedings of Scopus. The Web of Science is also a search engine with rigorous criteria and indexes mainly journals. This engine is responsible for the Journal Citation Report (JCR)\footnote{\url{https://jcr.clarivate.com/jcr/home}}, an evaluation metric of the impact of journals and publishers. Finally, this review includes gray literature (e.g., book chapters) that may not be indexed in academic search engines.

\subsection{Eligibility Criteria and Search Strategy} \label{eligibility-criteria}

The authors used a snowballing search strategy to identify synonyms to complement the search string. The query for outliers included terms such as \textit{outlier}, \textit{anomaly}, \textit{abnormality}, \textit{discordant}, \textit{deviant}, \textit{exception}, \textit{peculiarity}, \textit{aberration} and \textit{surprise}. Moreover, it considered \textit{clustering} and \textit{topic modeling}, including US and UK English particularities. There were also expressions concerning text data, which were \textit{natural language processing}, \textit{NLP}, and \textit{text mining}. In the context of initialization, the string contained \textit{initialization}, and analogously included expressions for clustering and NLP like in the outlier search string. There is a data extraction protocol to reduce variability, which is presented in Subsection \ref{subsec:data-collection}.

The systematic review included all works retrieved in ACLweb, Scopus, and Web of Science. The exclusion criteria consisted of retrieved proceedings with full editions, duplicate works amongst search engines, and duplicate works between initialization and outlier search strings. Besides, there were included published works from 2011 up to November 2021 written in English, whose scope involved topic modeling, clustering or related algorithms, text data, outliers, or initialization (whether the work proposed specific algorithms that tackle these issues or defined them). The authors repeated the search to double-check the retrieved works. Some works were not retrieved in the second attempt, but they met the eligibility criteria. Therefore, they were included in this literature review. The articles whose download was unavailable with our licenses (the institutional one and another of a national research foundation) without available preprint versions were excluded. In addition, the authors used the same search strategy, regardless of the search engine.

\begin{figure}
    \centering
    \includegraphics[width=\textwidth]{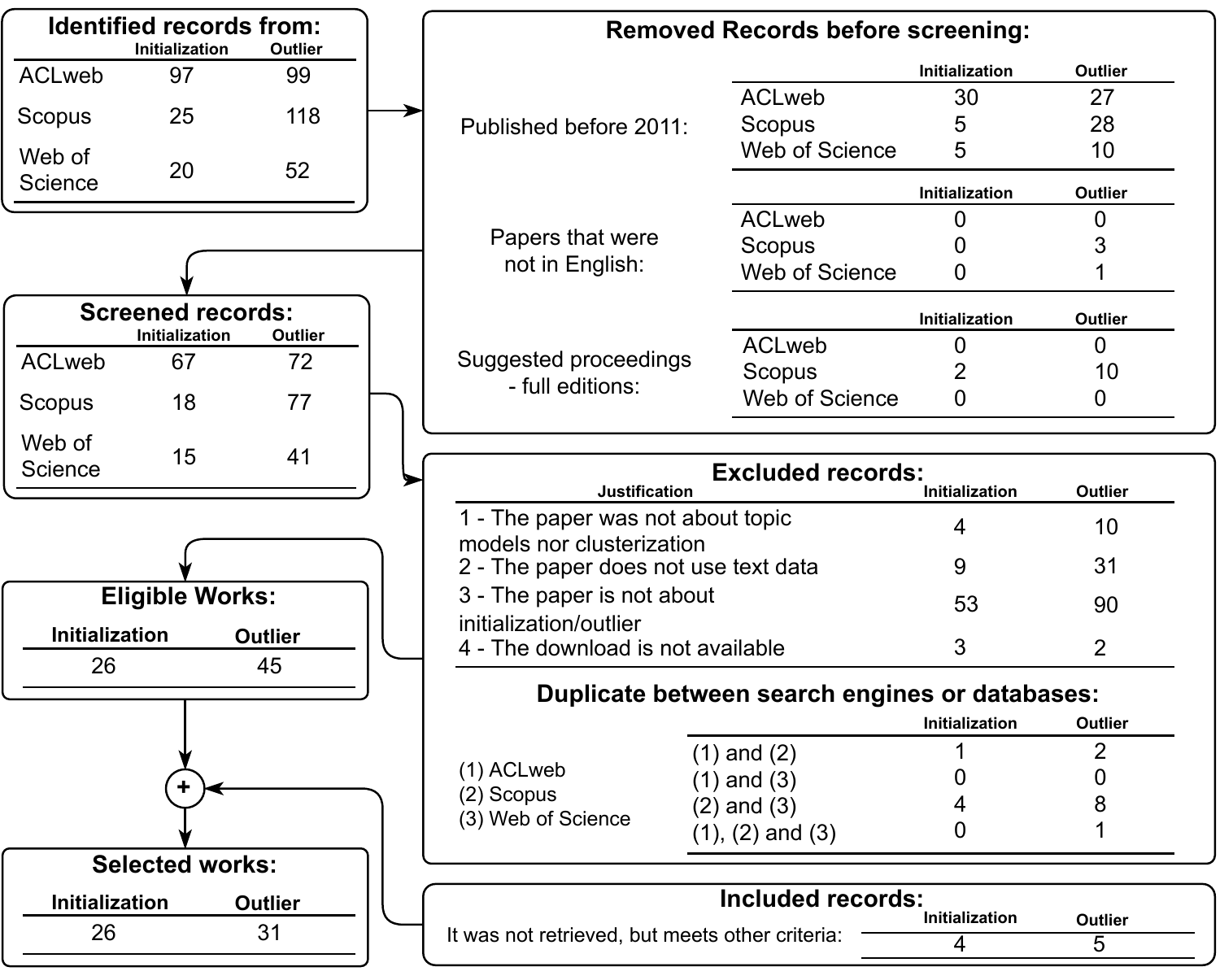}
    \caption{PRISMA flowchart for reviewed works.}
    \label{fig:prisma-data}
\end{figure}

The researchers included \textit{nine} extra literature works (e.g., book chapters, surveys), even though they would be older than the 10-year time frame. Each screened paper was revised, at first, by one reviewer, in order to exclude the ones that were not eligible. The works that were ambiguous and the eligible ones were independently scrutinized by three reviewers, who did not have access to the others' spreadsheets. After selecting, we calculated the kappa ($\kappa$) coefficient (see Equation \ref{eq:kappa_coef}). First, based on initialization, only works exclusively about this topic were considered, and the same occurred with the case of outliers. Nevertheless, we calculated the global $\kappa$ with all the selected works and all the eligible works, regardless of the groups they belonged to. The $\kappa$ for initialization was 68.00\%, whereas the $\kappa$ outliers was 48.94\%. The global $\kappa$ coefficient was 55.56\%.
\begin{equation}
    \label{eq:kappa_coef}
    \kappa = \frac{
            \text{Selected works based on unanimity}
        }{
            \text{Eligible works}
        } * 100\%
\end{equation}
Despite the disagreement, the divergent decisions were analyzed individually by the reviewers' panel. Finally, Figure \ref{fig:prisma-data} presents the diagram with the performed steps to screen the records and to select the eligible works about outliers and initialization. The analysis of three reviewers reduced the number of eligible works, and the sum of the remaining works and the included records are the total of selected works.

\subsection{Data Collection and Partiality Issues}\label{subsec:data-collection}

The authors predefined a list of questions and topics to guide the data extraction and to mitigate the partiality. From the selected works, 30\% were scrutinized by the reviewers to settle the data extraction process. The remaining 70\% were divided equally amongst the reviewers.

The data were sought concerning: corpora, learning approach (unsupervised, supervised, or semi-supervised), complementary learning methods (e.g., active learning, reinforcement learning), clustering and topic modeling algorithms (including the number of clusters, geometry, and group balance), preprocessing methods, text vectors, metrics (performance, concordance, similarity, and dissimilarity), initialization and outlier concepts, user interaction, computational implementation, visualization tools, outlier detection and initialization algorithms, objectives of the works, findings, advantages, drawbacks and research opportunities. Furthermore, the data extraction process included all the selected works. Each reviewer had tables to fill in the information about each paper. Besides, there could be appended remarks or citations that the reviewer considered necessary for each reviewed work. The collected data are presented throughout this review paper, summarized by the authors. Finally, the reviewers included ``not mentioned'' to the corresponding gap in the data extraction tables to enhance traceability whenever there was missing information.

\subsection{Synthesis of Reviewed Algorithms}\label{subsec:synthesis-reviewed}

This subsection synthesizes relevant reviewed work characteristics. They are categorized from different perspectives, aiming to improve the understanding of which techniques and algorithms the topic modeling research area has been using recently, also emphasizing outlier handling and initialization strategies for one or more components in a typical topic modeling pipeline.

Clustering algorithms are commonly found in the core of topic modeling pipelines, alongside proper probabilistic models designed for such task and factorization algorithms. The K-means algorithm is a popular baseline algorithm to evaluate more sophisticated algorithms. Table \ref{table:clustering} summarizes the common clustering or topic modeling algorithms among the reviewed works.
\begin{table}
    \caption{Reviewed works categorized by clustering or topic modeling algorithms they consider in their experiments. Some works may appear in multiple categories. ``DBSCAN'' stands for ``Density-based Spatial Clustering of Applications with Noise'', `PAM'' for ``Partition Around Medoids'', ``LDA'' for ``Latent Dirichlet Allocation'', ``LSA/LSI'' for ``Latent Semantic Analysis/Indexing'', ``NMF'' for ``Non-negative Matrix Factorization'', and ``pLSA/pLSI'' is the probabilistic LSA/LSI.}
    \begin{tabular*}{\textwidth}{p{4.5cm}p{9cm}}
        \hline
        \textit{Clusterer/ Topic Model} & \textit{Works}\\
        \hline \hline
        DBSCAN & \cite{Deepak2018, Suligoj2015, Thaiprayoon2020} \\
        Hierarchical & \cite{Bandi2020, Schubert2014, Ciosici2020,Sharuee2018, Kim2017, Babur2016, Thaiprayoon2020, Mehta2012}\\
        K-means & \cite{Buatoom2020, rakib2020enhancement, Pu2017, Sharuee2018, Vardasbi2019, Li2017, Liu2015, Zhu2013, Salgado2016, Suligoj2015, Thaiprayoon2020, Casalino2017, Onan2017, Shi2011, Babur2016, Hassani2021, Morup2012}\\
        K-medoids (PAM) & \cite{Onan2017, Salgado2016, Sharuee2018, DeAmorim2013}\\
        LDA & \cite{Peng2014, Bhatia2018, Onan2017, Zhuang2017a, Xue2011, Sholikah2020}\\
        LSA/LSI & \cite{Wang2019b, Zhu2013, Thaiprayoon2020}\\
        NMF & \cite{Liu2015, Casalino2017, Morup2012, Febrissy2020}\\
        pLSA/pLSI & \cite{Veselova2020, Xue2011} \\
        Others & \cite{Kim2016, Bandi2020, Wecker2020, Gubiani2017, P2016, Bhatia2018, Zhu2013, Zhuang2017a, Lazhar2019, Morup2012, Deepak2018, Hassani2021}\\
        \hline
    \end{tabular*}
    \label{table:clustering}
\end{table}

Some algorithms popularly used in topic modeling are sensitive to initial conditions, such as K-means \cite{Thaiprayoon2020, Zhu2013} or NMF \cite{Febrissy2020}. Some reviewed works run multiple random initialization, selecting the best at the end according to particular criteria, or proposing algorithms to initialize their clustering models in an informed way. All revised works that developed their initialization strategies are summarized in Table \ref{table:initialization}. For the sake of brevity, models initialized from publicly available pretrained weights were omitted from this table but included in Table \ref{table:embedding-methods}.

\begin{table}
    \caption{Reviewed works grouped by initialization. Some works may appear in multiple categories. All categories may include randomness but differ by using a more complex or informed initialization procedure. ``Multi-stage heuristics'' category comprises works that explicitly described their initialization procedures in two or more stages.}
    \begin{tabular*}{\textwidth}{p{4.5cm}p{9cm}}
        \hline
        \textit{Initialization category} & \textit{Works}\\
        \hline \hline
        Multi-stage heuristics & \cite{DeAmorim2013, Xue2011,Thaiprayoon2020, Zhu2013,Morup2012, Onan2017,Schubert2021, P2016}\\
        Random (multiple seeds) & \cite{Suligoj2015, Bhatia2018,Febrissy2020, Morup2012,Vardasbi2019} \\
        Random (single seed) & \cite{Casalino2017, Li2017} \\
        Others & \cite{Casalino2017, Febrissy2020,Hassani2021, Shi2011, Sholikah2020} \\
        \hline
    \end{tabular*}
    \label{table:initialization}
\end{table}

From all the 57 works considered, 31 (54\%) handled outliers during the experiments, whereas only 16 (52\%) of them clearly defined what an ``outlier'' is. Those definitions vary considerably from work to work, from more general ones (e.g., ``instances in low-density regions'') to definitions specific to text data (e.g., ``topic that does not belong to the given text document''). The strategies that adopted the outlier handling range from pure similarity analysis (e.g., cosine similarity distribution) to density-based algorithms (e.g., Local Outlier Factor (LOF) \cite{Breunig2000}). Five works proposed original algorithms robust to outliers by design so that the outliers would not compromise the results. The outlier handling strategies for each work can be seen in Table \ref{table:outlier}.
\begin{table}
    \caption{Outlier handling among the reviewed works. Some works may fit into multiple categories.}
    \begin{tabular*}{\textwidth}{p{4.5cm}p{9cm}}
        \hline
        \textit{Outlier handling strategy} & \textit{Works}\\
        \hline \hline
        Density-based & \cite{Suligoj2015, Consoli2021, Lazhar2019,Kim2017,Gubiani2017,Mehta2012, Schutze2011, Seo2020,Walkowiak2019, Shi2011} \\
        Robust to outliers by design & \cite{Kannan2017, Liu2015,Shi2011, Thaiprayoon2020,Zhuang2017a} \\
        Similarity-based & \cite{Camacho-Collados2016, Santus2018,Mohotti2020, Wang2019b} \\
        Others & \cite{Bhatia2018, Peng2014,rakib2020enhancement, Schubert2014} \\
        \hline
    \end{tabular*}
    \label{table:outlier}
\end{table}

Working with text data requires embedding of words, sentences, or documents into a vector space model as numerical vectors to enable optimization over the data. Among the reviewed works, Term Frequency-Inverse Document Frequency (TF-IDF) \cite{Luhn1957, Jones72astatistical, Salton1983IntroductionTM} was the most common vectorization technique, used by 17 (30\%) works, followed by 11 (19\%) works using Bag-of-words (BOW) or Bag-of-$n$-grams, and word2vec \cite{mikolov2013efficient} and \cite{mikolov2013distributed} was found in 9 (16\%) works. Other embedding strategies used in two or more reviewed works were GloVe (Global Vectors) \cite{pennington2014glove}, doc2vec \cite{le2014distributed}, Latent Dirichlet Allocation (LDA) \cite{Blei2003} or Non-negative Matrix Factorization (NMF) \cite{Paatero1994, Daniel1999} (both as intermediary representations) and fastText \cite{bojanowski2017enriching, joulin2016bag, mikolov2018advances}. Table \ref{table:embedding-methods} summarizes these findings.
\begin{table}
    \caption{Identified embedding methods for all reviewed works. Some works may appear in multiple categories as they experimented with different embedding methods. ``LDA'' and ``NMF'' works listed here used these algorithms to generate intermediary representations in the topic modeling pipeline, even though they require preliminary data embedding.}
    \begin{tabular*}{\textwidth}{p{4.5cm}p{9cm}}
        \hline
        \textit{Embedding method} & \textit{Works}\\
        \hline \hline
        BOW/Bag-of-$n$-grams & \cite{Sharuee2018, Suligoj2015,Avros2018, Kannan2017,Kim2017, Liu2015,Deepak2018, Peng2014, Babur2016,Bianchi2021, Li2017}\\
        doc2vec & \cite{Lazhar2019, Seo2020, Walkowiak2019} \\
        fastText & \cite{Aldarmaki2018, Walkowiak2019} \\
        GloVe & \cite{Bandi2020, Bhatia2018,Mohotti2020, Santus2018,Camacho-Collados2016}\\
        LDA & \cite{Peng2014, Onan2017, Vardasbi2019,Veselova2020, peinelt-etal-2020-tbert} \\
        TF-IDF & \cite{Sharuee2018, Suligoj2015, Casalino2017, Febrissy2020, Mohotti2020, Morup2012, P2016, rakib2020enhancement, Schubert2014, Seo2020, Shi2011, Zhuang2017a, Wang2019b, Zhu2013, Buatoom2020, Hassani2021}\\
        Transformer encoder & \cite{grootendorst2020bertopic, peinelt-etal-2020-tbert,Bianchi2021} \\ 
        word2vec & \cite{Camacho-Collados2016, Cassady2017,Li2017, Pu2017, Santus2018,Sholikah2020, Walkowiak2019,Wecker2020, Zhuang2017a} \\
        Others & \cite{Bandi2020, Consoli2021,Gubiani2017, Mehta2012,Thaiprayoon2020,Xue2011,Zhu2013, Zhuang2017a}\\
        \hline
    \end{tabular*}
    \label{table:embedding-methods}
\end{table}

From the 40 (70\%) reviewed works that reported empirical results, the most common metrics were accuracy, precision, recall, F1-score (harmonic mean of precision and recall), and Normalized Mutual Information (NMI) \cite{Kvalseth1987}. When NMI is normalized by the arithmetic mean of cluster assignment entropy from two distinct sources, it is equivalent to the V-measure (harmonic mean of cluster assignment completeness and homogeneity) \cite{Rosenberg2007}. Adjusted Mutual Information (AMI) \cite{Nguyen2010} is an NMI variant where the measure expected value for random agreement is accounted for, a useful property when the ratio of instance count to the number of clusters is small. Note that both NMI and AMI correspond to a family of related metrics as the normalization strategy varies (see \cite{Kvalseth1987, Nguyen2010}). Hence comparing reported results from distinct works needs to be done carefully.

\begin{figure}
    \centering
    \begin{subfigure}{0.32\textwidth}
        \includegraphics[width=\textwidth]{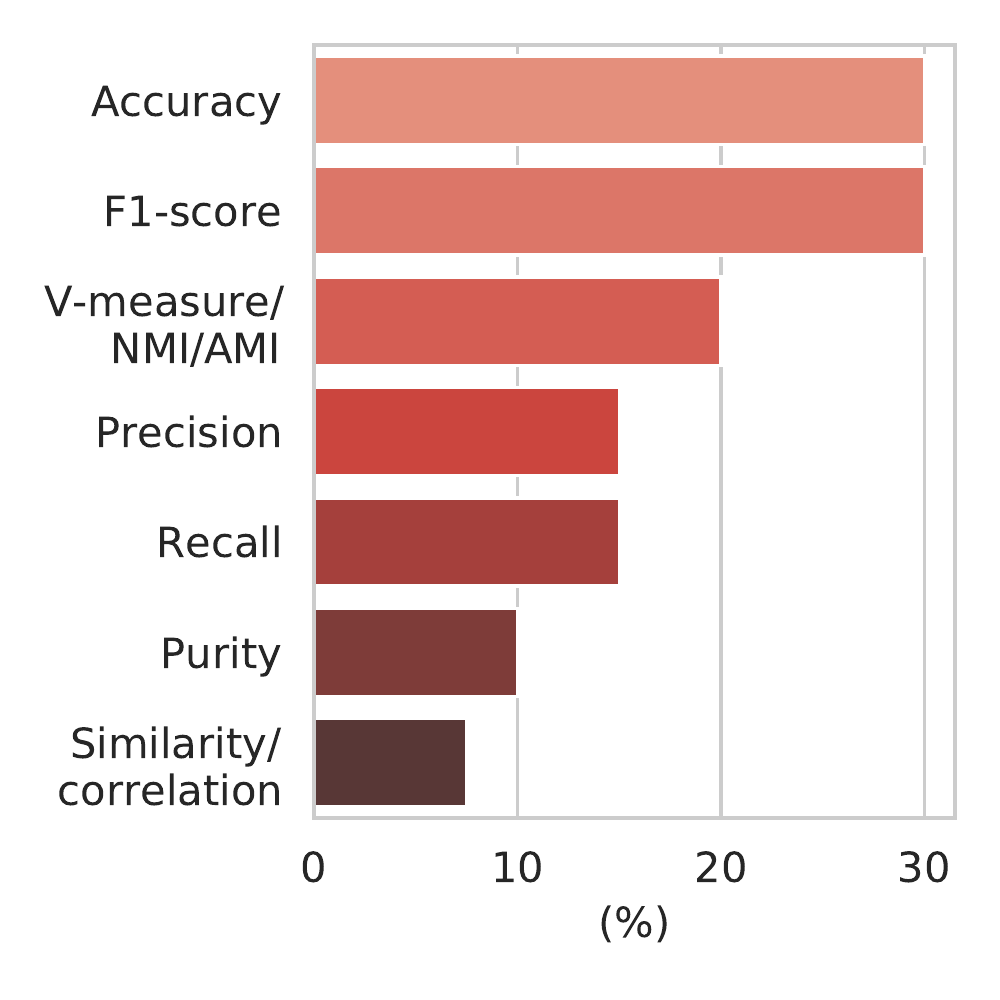}
        \caption{Evaluation metrics.}
        \label{subfig:barplot-metrics}
    \end{subfigure}
    \hspace*{\fill}
    \begin{subfigure}{0.32\textwidth}
        \includegraphics[width=\textwidth]{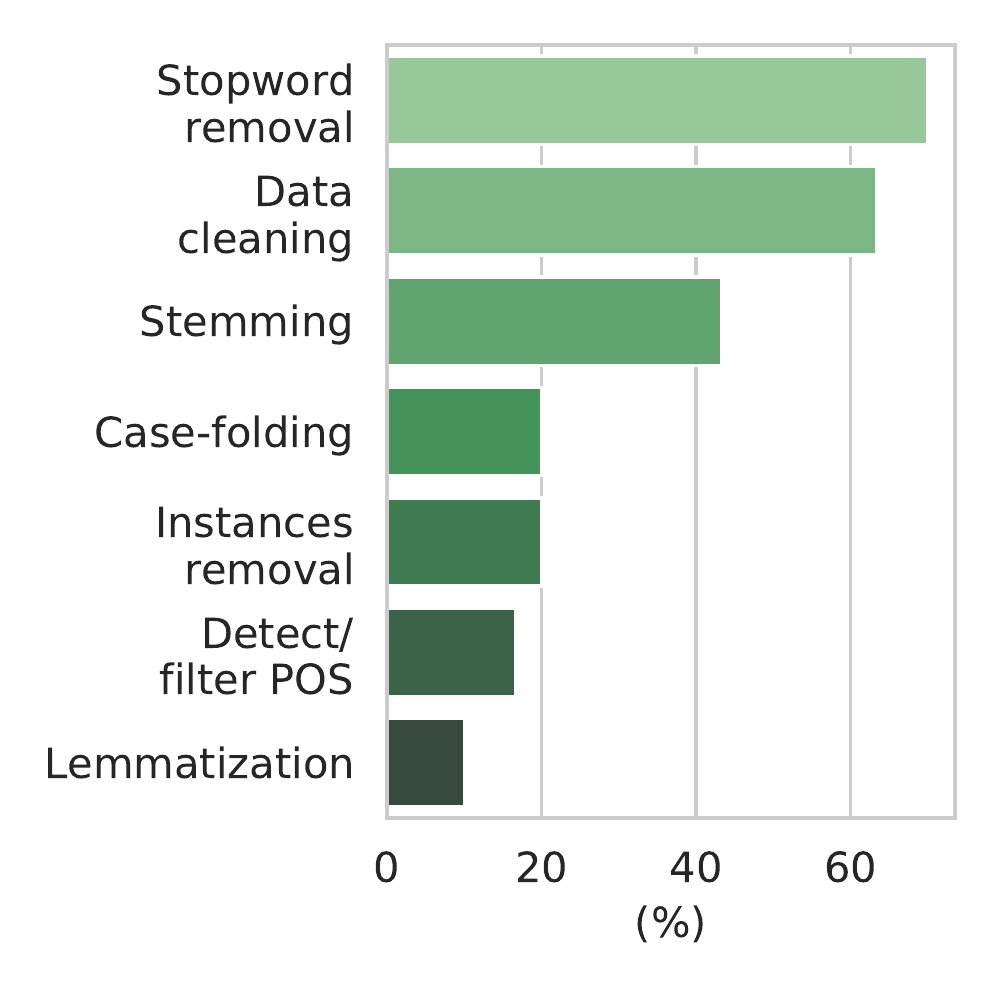}
        \caption{Preprocessing methods.}
        \label{subfig:barplot-preprocessing}
    \end{subfigure}
    \hspace*{\fill}
    \begin{subfigure}{0.32\textwidth}
        \includegraphics[width=\textwidth]{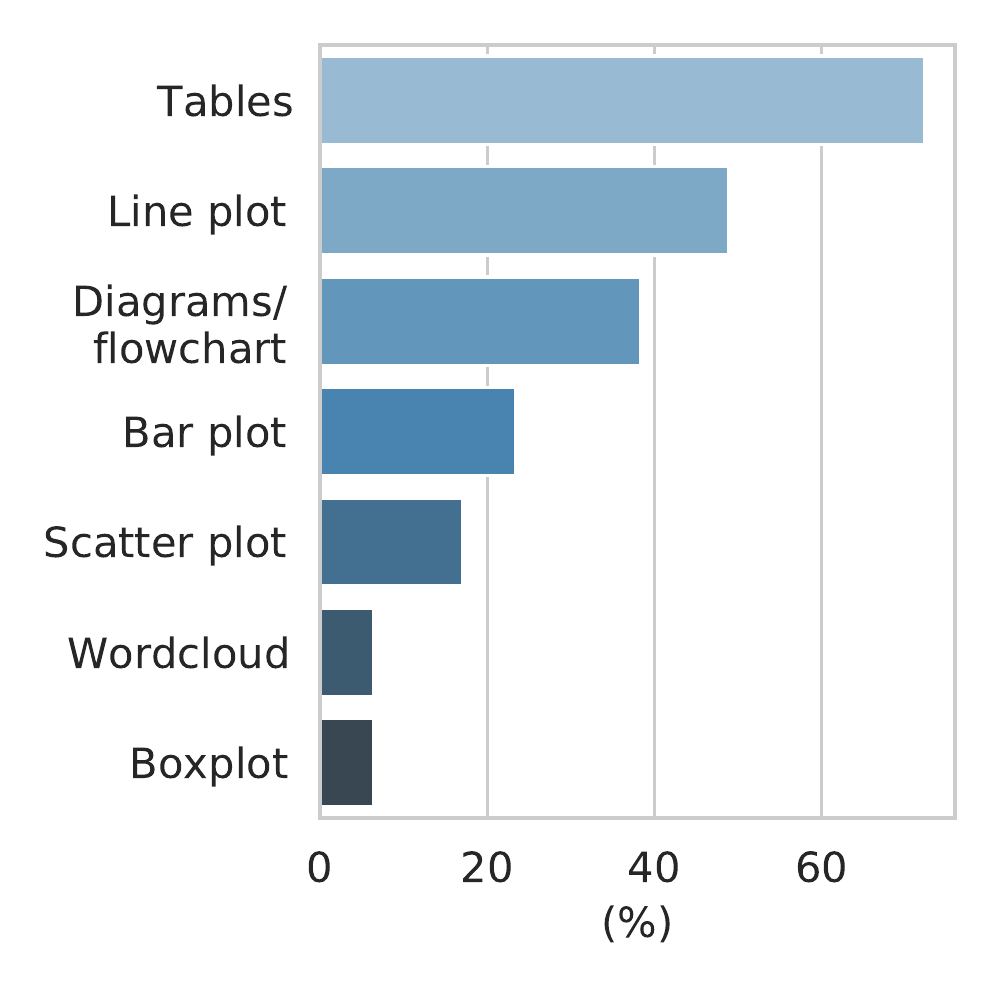}
        \caption{Visualization tools.}
        \label{subfig:barplot-visualization}
    \end{subfigure}
    \caption{Percentage of metrics (Figure \ref{subfig:barplot-metrics}), preprocessing methods (Figure \ref{subfig:barplot-preprocessing}), and visualization tools (Figure \ref{subfig:barplot-visualization}) from the reviewed works that reported these data.}
    \label{fig:metadata-barplots}
\end{figure}

The accuracy and F1-score were used mainly for supervised proof of concept studies of unsupervised algorithms. Only 4 reviewed works used statistical tests to validate their results \cite{Kim2017, Zhu2013, Hassani2021, Seo2020}. From the 30 (53\%) works that developed their data preprocessing steps, the most common choices (besides text tokenization) were text cleaning (e.g., removing unwanted symbols), stopword removal, stemming, and case-folding. When analyzing their results, tables, line graphs and flowcharts were predominant. Figure \ref{fig:metadata-barplots} summarizes these findings.

\begin{table}
    \caption{Programming languages and reviewed works with available implementations. Some publications provided implementations in more than one language.}
    \begin{tabular*}{\textwidth}{p{4.5cm}p{9cm}}
        \hline
        \textit{Programming Language} & \textit{Works}\\
        \hline \hline
        C/C++ & \cite{Ciosici2020} \\
        Java & \cite{Schubert2021} \\
        Matlab & \cite{Morup2012, Salgado2016, Kannan2017} \\
        Python & \cite{Li2017, Camacho-Collados2016,Salgado2016, Santus2018, Ciosici2020,Wecker2020, Bandi2020, rakib2020enhancement,grootendorst2020bertopic, peinelt-etal-2020-tbert,Schubert2021, Bianchi2021}\\
        R & \cite{Salgado2016, Schubert2021} \\
        \hline
    \end{tabular*}
    \label{table:prog-lang-implement}
\end{table}

Table \ref{table:prog-lang-implement} summarizes the works with available implementations. Only 14 works (from 57 reviewed works, which included the gray literature and papers that could not be retrieved but met the criteria) provided their implementations, hindering the research area efforts by reducing reproducibility.

The machine learning paradigms used by the reviewed works were unsupervised as the predominant paradigm with 49 (86\%) works, followed by supervised learning with 6 (11\%) works, and, finally, semi-supervised learning with 3 (5\%), considering that few works used multiple paradigms.

\section{Outlier Detection Benchmarks}
\label{sec:existing-benchmarks}

In 2016, \cite{Campos2016evaluationofunsupervised} surveyed general-purpose unsupervised algorithms based on k-nearest neighbors to create a benchmark. \cite{Goldstein2016comparativeevaluation} also addressed unsupervised anomaly detection, dividing the algorithms into nearest-neighbor based, clustering-based, statistical, subspace-based, and classifier-based/other. \cite{emmot2015metaanalysis} presented a meta-analysis of outlier detection algorithms and the existing benchmarks. \cite{DOMINGUES2018comparativeevaluation} and \cite{Steinbuss2021benchmarkingunsupervised} also developed benchmarks to the unsupervised outlier detection problem. In 2021, \cite{Ruff2021unifyingreview} surveyed shallow and deep outlier detection algorithms, dividing them into classification-based, probabilistic, reconstruction-based, and distance-based. Next, \cite{SoenenHyperparameterTuning2021} analyzed the problem of how to fine-tune and select the hyperparameters of unsupervised outlier detection algorithms. Finally, in 2022, \cite{Han2022ADBench} proposed a new benchmark concerning anomaly detection, with the perspective of the levels of supervision, the contamination of data, and the types of anomaly (e.g., local, global, group anomalies, and dependency anomalies), using the default hyperparameters.

\cite{Goldstein2016comparativeevaluation} and \cite{Ruff2021unifyingreview} mentioned the problem of the curse of dimensionality when dealing with high-dimensional data, which is a frequent concern in NLP. The eight benchmarks analyzed in this survey were focused on tabular data. In 2022, \cite{Han2022ADBench} is the first work to consider NLP datasets (4 amongst the 55). These text data were embedded with Bidirectional Encoder Representations from
Transformers (BERT) \cite{Devlin2019}, extracting the [CLS] token.  
\cite{emmot2015metaanalysis}, \cite{Steinbuss2021benchmarkingunsupervised}, \cite{Ruff2021unifyingreview} and  \cite{Han2022ADBench} used synthetic data. The strategy of downsampling was used to create outliers in works like \cite{Campos2016evaluationofunsupervised, Han2022ADBench}. \cite{Steinbuss2021benchmarkingunsupervised} provided an in-depth analysis concerning the synthetic datasets. 

With the type of the outlier known in advance \cite{Campos2016evaluationofunsupervised, Goldstein2016comparativeevaluation, Han2022ADBench} or the context of the dataset \cite{emmot2015metaanalysis}, it is possible to define well-tailored decision-making concerning the outlier detection algorithms \cite{Campos2016evaluationofunsupervised}. Besides, the data properties (e.g., dimension, number of instances, scale, hierarchy, contamination, context, and so forth), domain, and application sustain assumptions about the choice of the algorithms \cite{Ruff2021unifyingreview}. \cite{SoenenHyperparameterTuning2021} advocated that it is not possible to compare performance metrics when some papers use the default settings, whereas others set different hyperparameters. Furthermore, the fine-tuning influenced the algorithm performance, and some anomaly detection algorithms cannot be tuned due to the computational overhead \cite{SoenenHyperparameterTuning2021}. In addition, datasets with only 100 labeled instances with anomalies (called validation set) led to a performance gain \cite{SoenenHyperparameterTuning2021}. According to \cite{Han2022ADBench}, the use of feature selection techniques can mitigate the influence of noisy data.

During the evaluation, the benchmarks used as their performance metrics the Precision at n (P@n) \cite{Campos2016evaluationofunsupervised,  Ruff2021unifyingreview}, Adjusted P@n \cite{Campos2016evaluationofunsupervised}, Average Precision \cite{Campos2016evaluationofunsupervised, emmot2015metaanalysis, DOMINGUES2018comparativeevaluation, Ruff2021unifyingreview}, Area Under the Curve of the Receiver Operating Characteristic Curve (ROC AUC) \cite{Campos2016evaluationofunsupervised,  Goldstein2016comparativeevaluation, emmot2015metaanalysis, DOMINGUES2018comparativeevaluation, Steinbuss2021benchmarkingunsupervised, Ruff2021unifyingreview, SoenenHyperparameterTuning2021, Han2022ADBench}, Area Under the Precision-Recall Curve (PR AUC) \cite{DOMINGUES2018comparativeevaluation, Steinbuss2021benchmarkingunsupervised, Ruff2021unifyingreview, Han2022ADBench}, and Average AUC \cite{SoenenHyperparameterTuning2021}. According to \cite{Ruff2021unifyingreview}, the PR AUC is a better performance metric than ROC AUC, since ROC AUC can be overly optimistic. Finally, \cite{Campos2016evaluationofunsupervised,  Steinbuss2021benchmarkingunsupervised, Han2022ADBench} concluded that, with their experiments, no anomaly detection algorithm outperformed the others regardless of the scenario, type of outlier, or dataset characteristics.

\section{Outlier Detection and Handling}
\label{sec:results-outlier-handling}

The exact definition of ``outlier'' can be difficult because it highly depends on the task context \cite{Chandola2009, Wang2019c}. For instance, in the context of Hierarchical Agglomerative Clustering (HAC), while \cite[p.382]{Manning2008} claims that complete-linkage is ``sensitive to outliers'', in \cite[p.558]{Tan2018} and \cite[p.283--284]{Steinbach2003} the complete-linkage is said to be ``more robust to outliers'' while single-linkage is sensitive to it. In \cite[p.533]{Timm2002}, the single-linkage is again said to be ``robust to outliers'', but sensitive to ``errors of measurement''. Since the single-linkage is more susceptible to the chaining effect, and the complete-linkage uses the maximum pairwise distance between clusters, both can have undesired effects due to data points in low-density regions. However, they do not share the same problem in a general sense. Hence, while the descriptions of these works use the same keyword ``outlier'', they do not describe the same issue. Some authors claim that other common names for outliers in the literature are \textit{novelty}, \textit{abnormalities}, \textit{discordants}, \textit{deviants} and \textit{anomalies} \cite{Aggarwal2013} and potentially many others, but whether all these terms actually refer to the same phenomena is not consensual \cite{Carreo2019}. These issues highlight the importance of a clear definition of what ``outlier'' is in the context of each separated work.

In the context of text data, outliers may be defined in word, sentence, or document level. Concerning topic modeling, outliers can be defined in the topic level, such as underrepresented topics \cite{Veselova2020} or topics that expand too fast \cite{Schubert2014}. From the reviewed works, 16 (52\%) works defined what an ``outlier'' means in the context of their experiments, from 31 total works that handled outliers. The majority of these works defined outliers in the document level. Documents that deviate from corpus distribution in a perceptive manner are a common definition pattern. Just a few works defined outlier in the word level \cite{Camacho-Collados2016, Santus2018, Peng2014} and in the topic level \cite{Schubert2014, Bhatia2018, Veselova2020}.

For the rest of this section, we will present a few techniques and algorithms regarding outlier handling, detection, or removal found in reviewed works.

Either similarity or density-based outlier detection methods need to compare embeddings. One common choice of similarity function is the cosine similarity between a pair of embeddings $a$ and $b$ \cite{Camacho-Collados2016, Wang2019b, Shi2011, Suligoj2015, Consoli2021, Gubiani2017}, as shown in Equation \ref{eq:cosine}.
\begin{equation}
    \label{eq:cosine}
    \similarity(a, b) = \cos(a, b) = 
    \frac{
        a^{T} b
    }{
        \left\Vert a \right\Vert_{2}
        \left\Vert b \right\Vert_{2}
    }
    =
    \frac{
        \sum_{i}^{} a_{i}b_{i}
    }{
        \sqrt{
            \sum_{i}a_{i}^2
            \cdot
            \sum_{i}b_{i}^2
        }
    }
\end{equation}
Based on \cite{Santus2016a} and \cite{Santus2016b}, in the reviewed work \cite{Santus2018}, a rank-based alternative (called APSync) for cosine similarity was proposed. It was shown to consistently outperform cosine similarity in outlier detection experiments. The APSynP definition is shown in Equation \ref{eq:rank-similarity}, where $\rank(v_{i}; v)$ denotes the index that $v_{i}$ falls when entries in vector $v$ are sorted decreasingly, and $p \in [0, 1]$ is a hyperparameter and fixed as $p=0.1$ in the paper experiments based on empirical validation.
\begin{equation}
    \label{eq:rank-similarity}
    \similarity(a, b; p) =
    \text{APSynP}(a, b; p) =
    \sum_{i}
    \frac{2}{\rank(a_{i}; a)^{p} + \rank(b_{i}; b)^{p}}
\end{equation}
The reviewed work \cite{Wang2019b} proposes to use Angle-Based Outlier Factor (ABOF) \cite{Kriegel2008}, which analyzes the variance between the angles of documents in the embedded corpus to identify ``outlier documents'', or documents consistently farther away from all other documents. The hypothesis behind this method is that, for a sufficiently farther away observer (an outlier), every other document is next to each other, and hence the variance of the angles between the observer and every other documents is small. This idea is formalized in Equation \ref{eq:abof} which calculates the ABOF for a single document $d$, where VAR denotes the empirical variance.
\begin{equation}
    \label{eq:abof}
    \text{ABOF}(d; D) = \mathop{VAR}_{i,j=1,\ldots,\lvert D\lvert}
    \left[
    \frac{
        (d - d_i)^{T}
        (d - d_j)
    }{
        \diffnorm{d}{d_i}_{2}^{2}
        \diffnorm{d}{d_j}_{2}^{2}
    }
    \right]
\end{equation}
Clustering is commonly at the core of topic modeling algorithms. Thus, it may be relevant to choose a clustering algorithm that is insensitive to outliers or can handle unusual observations during the execution. Classical clustering algorithms such as PAM, Clustering LARge Applications (CLARA), and Clustering Large Applications based on RANdomized Search (CLARANS) are potentially robust to outliers (if a proper similarity function is chosen), while DBSCAN and Hierarchical DBSCAN (HDBSCAN) are able to handle outliers naturally, although in this case outliers are often considered undesired noise instances in the data. Since these algorithms are employed to cluster documents or words, the outlier detection is often restricted to word and document levels and not to the topic level.

In \cite{Suligoj2015}, classical K-means and intra-cluster similarity were combined, where documents that have low similarity with their respective centroids are considered outliers. In \cite{Gubiani2017}, a similar approach was taken when documents were first clustered in two clusters and afterward may be reassigned to the other cluster grounded on which cluster the document has higher cosine similarity.

In \cite{Schubert2014}, \textit{rapidly emerging topics} are considered outliers. This approach differs from the majority of other works since it is applied to data streams. For each topic $w$ in a vocabulary V, represented as word $n$-grams, its significance score is calculated as shown in Equation \ref{eq:significance}, where $f : V \rightarrow \mathbb{R}_{+}$ maps each topic to its frequency, $\mu, \sigma^{2} : V \times \mathbb{N} \rightarrow \mathbb{R}_{+}$ are the moving average and variance of the topic frequency respectively, $\eta > 0$ is a learning rate for both moving statistics, and $\beta > 0$ is a bias term that avoids division by zero and also smooths the formula to be more robust to rare terms.
\begin{equation}
    \label{eq:significance}
    \text{significance}(w, t; f, \eta, \beta) =
    \frac{
        f(w) - \max\left\{\mu(f(w), t; \eta), \beta\right\}
    }{
        \sqrt{\sigma^{2}(f(w), t; \eta)} + \beta
    }
\end{equation}
If a topic $w$ has a significance level above a threshold $T = \mu + s\sigma^2$ for some predetermined $s > 0$, then $w$ is considered an outlier (rapidly emerging topic). As it handles data streams, the significance level is checked periodically.

In \cite{Lazhar2019}, Fuzzy K-means was used to detect outlier documents. More precisely, the Fuzzy K-means (see the objective function in Equation \ref{eq:fuzzy-k-means}) is run in the corpora embedded with doc2vec, and documents close to the cluster decision boundaries are considered potential outliers and inserted into a subset $O$. Then, the Fuzzy K-means is rerun into $O$, and every outlier candidate that sufficiently increases the objective function is declared an outlier by the algorithm.

In \cite{Zhuang2017a}, a probabilistic generative algorithm is proposed for topic modeling similar to LDA called Embedded von Mises-Fisher Allocation (EvMFA), by using embeddings from word2vec as input to the generating process, therefore generating documents as a bag-of-normalized-embeddings instead of a bag-of-words similar to LDA by following a von Mises-Fisher distribution (the Gaussian distribution analog in directional statistics) instead of a categorical distribution for each word. Equation \ref{eq:vmf-pdf} shows the probability density function for von Mises-Fisher distribution, where $i$ is the topic index, $d \in \mathbb{N}^{E}$ is a $E$-dimensional embedded document, $\kappa > 0$ is the concentration parameter (thus inversely proportional to the variance), $\mu_i \in \mathbb{R}^E$ is the mean direction for topic $i$ such that $\Vert \mu_i \Vert_{2} = 1$, and $I_{z}$ is the modified Bessel function of the first kind at order $z$.
\begin{equation}
    \label{eq:vmf-pdf}
    \pparam{d}{\kappa_i, \mu_i, E} =
    \frac{
        \kappa_i^{(E/2 - 1)}
    }{
        (2\pi)^{E/2}
        \cdot
        I_{(E/2 - 1)}(\kappa_i)
    }
    \exp(\kappa\cdot\cos(\mu_i, d))
\end{equation}
Similar algorithms were previously used in \cite{Reisinger2010, batmanghelich2016} also for topic modeling, but \cite{Zhuang2017a} focuses its application on outlier detection. First, it uses EvMFA to model the topic regions, or \textit{semantic focus} regions. Then, it filters out regions with low concentration (low estimated $\kappa$) grounded on a hyperparameter $\beta \in [0, 1]$, which corresponds to a cutoff to the semantic focus cumulative probability distribution. Lastly, it uses a criterion named Orthodox Quantile Outlierness (OQO) $\Omega(d; \theta)$ to compute the probability of whether a document is an outlier, based on how many \textit{orthodox tokens} compose it. Orthodox tokens are tokens that both are found close to semantic focus areas and also sufficiently specific to corpus $D$ against a background corpus $D_\text{bg}$. Equation \ref{eq:vMF-Q-outlier} shows the OQO criterion, where $q(\,\cdot\,; \theta)$ denotes the $\frac{1}{(1-\theta)}$-quantile function of Poisson-Binomial distribution, and $\omega_\text{orthodox}$ is a function that estimates the orthodox token count within document $d \in D$ against a background corpus $D_\text{bg}$.
\begin{equation}
    \label{eq:vMF-Q-outlier}
    \Omega(d; \theta, D, D_\text{bg}) = 1 - \frac{
        q(\omega_\text{orthodox}(d; D, D_\text{bg}); \theta) + 1
    }{
        \lvert d\lvert + 1
    }
\end{equation}
Other general-purpose outlier detection algorithms were also employed in the reviewed works: Local Outlier Factor (LOF) \cite{Breunig2000} was present in 3 works \cite{Kim2017, Seo2020, Walkowiak2019}, One-class Support Vector Machine (One-class SVM) \cite{Scholkopf1999} in 2 works \cite{Sholikah2020, Avros2018} and Isolation Forest \cite{Liu2009, Liu2012} used in a single work \cite{rakib2020enhancement}. All these algorithms are available in the scikit-learn library \cite{scikit-learn, sklearn_api}.

In \cite{Kannan2017}, an NMF variant ("Text Outliers using Nonnegative Matrix Factorization (TONMF)") was developed to detect outlier documents. The objective function resembles Equation \ref{eq:nmf}, but assumes that $D = L_{0} + Z_{0}$, where $Z_{0}$ is a matrix of outlier factors and $L_{0}$ is the true document embeddings generated by an underlying process. Thus, the new objective function is given by Equation \ref{eq:nmf-outlier}.

\begin{align}
    \begin{split}
        \label{eq:nmf-outlier}
        \min\limits_{W,H,Z} J(W, H, Z; D, \alpha, \beta) & = \min\limits_{W,H,Z} \tfrac{1}{2} \left\Vert D - WH - Z \right\Vert_{F}^{2}
        + \alpha\left\Vert Z \right\Vert_{(2,1)}
        + \beta\left\Vert H \right\Vert_{1}
    \end{split} \\
    \text{subject to}
        & \quad W, H \geq 0 \notag
\end{align}
In this equation, $\alpha \geq 0$ controls the sensitivity of the algorithm for outliers, $\beta \geq 0$ is a regularization factor to provide sparsity to the results, and $\left\Vert\,\cdot\,\right\Vert_{(p,q)}$ is a matrix-mixed norm defined in Equation \ref{eq:mixed-norm}. During the experiments, the authors noticed that $\beta$ has low influence in the results.
\begin{align}
    \label{eq:mixed-norm}
    \left\Vert Z \right\Vert_{(p,q)} =
    \left(
        \sum_{i}
        \left(
            \sum_{j}
            \lvert z_{(i,j)}\lvert^{p}
        \right)^\frac{q}{p}
    \right)^\frac{1}{q}
\end{align}
Unlike the original paper that defines $D$ so that each \textit{column} corresponds to a document, in the present work, $D$ is defined such that each document is represented by a \textit{row}, and hence the mixed-norm term in Equation \ref{eq:nmf-outlier} was adapted to reflect our convention. After the optimization procedure, the $\ell_{2}$-norm from the rows of the outlier matrix $Z$ are calculated, and documents with high scores are considered outliers by this particular method.

Another NMF variant called $\ell_{1}$-norm Symmetric Nonnegative Matrix Trifactorization ($\ell_{1}$ S-NMTF) was proposed in \cite{Liu2015} to perform NMF while effectively handling outlier documents during the optimization process, showing noticeable improvements over vanilla NMF and other algorithms in their experiments. The authors also provide an optimization algorithm. The drawback of this method is that it requires a particular representation of $D$.

The reviewed work \cite{Peng2014} pointed out that a small amount of data available leads to less coherent and interpretable topics. To cope with this, they used Regularized LDA \cite{Newman2011}, which also improves results for text with noise. It works by choosing specific priors based on statistics of external, sufficiently large corpora $D_\text{external}$, and plugging it into the traditional LDA. More precisely, a ``covariance'' matrix $C = WW^{T}$ is computed, where $W$ is a co-occurrence matrix of top-$k$ most current words from $D_\text{external}$ within a moving window through the documents. It was shown in the original paper to improve mutual information against human judgment for topic modeling and may decrease model perplexity slightly.

The reviewed work \cite{Bhatia2018} uses Convolutional Neural Networks (CNN) \cite{Fukushima1982, LeCun1989} algorithms combined with scores from BM25 \cite{Robertson2009} to detect unrelated topics to a given document, the \textit{intruder topic}, from a list of candidate topics that may belong to the document. Unlike most reviewed works, this one adopts the supervised learning paradigm for outlier topic detection.

\section{Initialization of Topic Modeling and Clustering Algorithms}
\label{sec:results-initialization}

Topic models often have components sensitive to initialization settings, which raises the question of what is the best initialization strategy. A straightforward strategy is to initialize each component multiple times with distinct random initialization and pick the setting that gives the best performance according to some metric(s) \cite{Suligoj2015, Bhatia2018, Febrissy2020, Morup2012}. While simple, this strategy may not be deployed for expensive pipelines. This section groups strategies from the reviewed works to initialize common components of topic modeling pipelines, which are designed to be either more robust to outliers or to improve the optimization convergence rate.

K-means++ \cite{Arthur2007} consists of the regular K-means with an initialization algorithm proposed to reduce the variability of the clustering result. It works by sampling the initial centroids iteratively while taking into account the centroids already selected under the hypothesis that centroids are better initialized far apart from each other. Choosing iteratively always the farthest possible data points from the current centroids is an approach sensitive to outliers. Thus, the K-means++ approach selects them randomly. The first centroid $c_{1}$ is taken arbitrarily. Then, each document $d$ in $D$ has a probability inversely proportional to the squared Euclidean distance to the closest centroid to become the next centroid, as shown in Equation \ref{eq:k-means++}, where $C$ is the set of centroids currently chosen.
\begin{equation}
    \label{eq:k-means++}
    p(d\lvert C;D) =
    \frac{
    \min\limits_{c \in C}
        \diffnorm{d}{c}_{2}^{2}
    }{
        \sum\limits_{z \in D}
        \min\limits_{c \in C}
        \diffnorm{z}{c}_{2}^{2}
    }
\end{equation}
This process is repeated $k - 1$ times until the requested number of centroids $k$ is gathered. After this initialization procedure, the regular K-means is run. From the reviewed works, 3 works used K-means++ in their experiments \cite{Vardasbi2019, Onan2017, Thaiprayoon2020} (corresponding to 18\% of works that used K-means).

The reviewed work \cite{Onan2017}, when initializing K-medoids, closely follows Equation \ref{eq:k-means++} by replacing centroids with medoids, an idea which was considered by \cite{Lijffijt2013}. However, \cite{Schubert2019, Schubert2021} show that such a strategy may impair PAM convergence rate and propose a PAM initialization named Linear Approximative BUILD (LAB), where it applies the original PAM initialization algorithm BUILD with subsamples of size in $\mathcal{O}(\lvert D\lvert^{\frac{1}{2}})$. For computationally cheaper versions of PAM, such as FasterPAM, the initialization procedure is often associated with the largest portion of computational cost rather than the algorithm optimization steps. Thus, cheaper initialization like K-means++ is preferred, even though it is less reliable than the costly but precise BUILD method \cite{Schubert2021}.

Even though K-means++ is a better initialization algorithm than random initialization, \cite{Bhaskara2019} argues that it is too sensitive to outliers. For this purpose, they propose the Thresholded K-means++ (T-kmeans++), which essentially sets an upper bound to the distance between each point and the currently chosen centroids before computing the K-means++ sample probabilities, thus reducing the probability of outlier centroids as shown in Equation \ref{eq:t-kmeans++}, where $T$ is a given threshold.
\begin{equation}
    \label{eq:t-kmeans++}
    p(d\lvert C;D,T) =
    \frac{
    \min\limits_{c \in C}
        \left\{
        \diffnorm{d}{c}_{2}^{2}, T
        \right\}
    }{
        \sum\limits_{z \in D}
        \min\limits_{c \in C}
        \left\{
        \diffnorm{z}{c}_{2}^{2}, T
        \right\}
    }
\end{equation}
In \cite{Shi2011}, a K-means initialization method called WIKTCM, which considers possible outliers, is proposed. The first centroid document is the document with the highest total similarity to all other documents. Then, the next centroid documents are chosen deterministically and iteratively grounded on a score proportional to the similarity score to all other remaining, non-outlier documents and also proportional to the dissimilarity with previously chosen centroids, as shown in Equation \ref{eq:wiktcm-init}, where $C$ denotes the set of previously chosen centroids, and $O$ denotes the set of currently identified outlier documents.
\begin{equation}
    \label{eq:wiktcm-init}
    c_\text{next} = \argmax\limits_{d \in (D - C - O)}
    \underbrace{
    \left[
        \frac{1}{\lvert D - C - O\lvert - 1}\sum_{z \in (D - C - O)}\cos(d, z)
    \right]
    }_\text{Average similarity with remaining documents}
    \cdot
    \underbrace{
    \left[
        \vphantom{
        \frac{1}{\lvert D - C - O\lvert - 1}\sum_{z \in (D - C - O)}\cos(d, z)
        }
        \sum_{z \in C}1 - \cos(d, z)
    \right]
    }_\text{Dissimilarity to centroids}
\end{equation}
The outliers are detected before every new centroid is chosen based on their average similarity score to all remaining documents (non-centroids and non-outliers). If a document has average similarity below a threshold, it is considered an outlier and thus removed from corpus.

Finally, in \cite{Vardasbi2019}, an influence graph-based initialization algorithm to the K-means was proposed, where this work claims improvements over the K-means++ initialization in its experiments using an algorithm of $\mathcal{O}(N\lvert E\lvert^{2}\lvert D\lvert^{-1})$ time complexity (where $N$ is the maximum number of iterations and $\lvert E\lvert$ the number of edges) called MMSE‑2, corresponding to the objective function shown in Equation \ref{eq:mmse-2}, the sum of squared errors of the approximated influence probabilities $p_{2}(d_i \ \rightarrow\ d_j)$ for every weighted graph edge $(d_i, d_j, w_{(i, j)}) \in E$, where $w_{(i,j)} \in [0, 1]$, $NN(d; D)$ denotes the set of neighbor indices to document $d \in D$ and the similarity function is arbitrary as long as it maps any pair of documents to the $[0, 1]$ range. This objective can be minimized using gradient descent.
\begin{equation}
    \label{eq:mmse-2}
    J(W; D, E) =
    \sum_{(d_i, d_j, \cdot) \in E}
    \Bigl(
    \underbrace{
        1 -
        \prod_{
            z \in \text{NN}(d_i; D) \cap \text{NN}(d_j; D) \\
        }(1 - w_{(i,z)}w_{(z, j)})
    }_{p_{2}(d_i\ \rightarrow\ d_j)}
    -
    \underbrace{
        \vphantom{
            \prod_{
                d_k \in \text{NN}(d_i; D) \cap \text{NN}(d_j; D) \\
            }(1 - w_{(i,k)}w_{(k, j)})
        }
        \similarity(d_i, d_j)
    }_{\in [0, 1]}
    \Bigr)^{2}
\end{equation}
There were efforts to address the NMF sensitiveness to the initial conditions \cite{Febrissy2020, Hassani2021, langville2014} in \cite{Boutsidis2008} whose propositions were three algorithms based on SVD (see Equation \ref{eq:svd}) to initialize matrices $W$ and $H$, namely Nonnegative Double Singular Value Decomposition (NNDSVD), NNDSVD Averaged (NNDSVDa), and NNDSVD Averaged Random (NNDSVDar). The first variant is suitable when sparseness in $W$ and $H$ is desired, and the last two make $W$ and $H$ denser with deterministic and random approaches, respectively. This algorithm is adopted in two reviewed works \cite{Hassani2021, Casalino2017}. Although these initialization algorithms were shown to improve the NMF convergence rate considerably, it is not clear if they perform any better than random initialization given enough iterations. These three initialization algorithms are available for the NMF implementation in the scikit-learn library \cite{scikit-learn, sklearn_api}.

The reviewed work \cite{Casalino2017} also experiments with two other initialization algorithms for the NMF proposed in \cite{langville2014} grounded on semi-informed NMF random initialization. Its experiments show that NNDSVD is the most accurate but also the slowest initialization algorithm. Although less accurate, the other two are significantly faster and also provide improvements over uninformed random initialization. This work also shows that the best NMF initialization may depend on which algorithm solves the NMF optimization problem and also on the data distribution. The implementation of these three algorithms is available at the Nimfa library \cite{Zitnik2012}.

For neural-based text embedding initialization, it is common to adopt publicly available pre-trained embeddings from surrogate tasks with huge, general-purpose corpora, thus leveraging natural language understanding from typically richer data than the available for the downstream task. The transfer learning enables these embeddings to be fine-tuned during the downstream task optimization. The algorithms word2vec, doc2vec, and fastText and their respective pre-trained embeddings for varying dimensions are available in Gensim library \cite{rehurek_lrec}, alongside all the datasets used for training. Pre-trained GloVe embeddings are also available online\footnote{\url{https://nlp.stanford.edu/projects/glove/} (for English language)}. Pre-trained transformers-based models (such as BERT and Sentence-BERT (SBERT)) are available in HuggingFace's hub \cite{Wolf2020}.

While embedding fine-tuning details is out of the scope of this present work, it is worth mentioning that often there is room for improving embedding quality for downstream tasks by fine-tuning to domain-specific data. For some distributed representations such as word2vec, doc2vec, and fastText, the same self-supervised algorithm can be used for such fine-tuning. For pre-trained embeddings originally from a supervised task such as SBERT, it is not obvious how to perform the fine-tuning since domain-specific labeled data may not be readily available. For such occasions, self-supervised contrastive learning algorithms are good candidates, such as Simple Constrastive Sentence Embedding (SimCSE) \cite{Gao2021} or Contrastive Tension \cite{Carlsson2021}.

\section{Topic Modeling as a Pipeline: End-to-End Topic Models}
\label{sec:end-to-end-topic-models}

Current end-to-end topic models may rely on transformer encoders integrated with another topic modeling, clustering, and embedding algorithms.

BERTopic \cite{grootendorst2020bertopic} is a topic modeling package that proposes creating topics from pre-trained SBERT document embeddings in a pipeline composed by dimensionality reduction with Uniform Manifold Approximation and Projection (UMAP) \cite{Mcinnes2020}, followed by document clustering with HDBSCAN, and then topic extraction using the so-called \textit{class-TF-IDF}, which consists of regular TF-IDF (see Equation \ref{eq:tf-idf}) applied separately for each identified cluster from HDBSCAN. Since HDBSCAN identifies candidate outlier documents, the BERTopic proposed pipeline takes them into consideration as well, however assuming outlier documents are noise and thus are not modeled.

tBERT \cite{peinelt-etal-2020-tbert} concatenates contextual embeddings from BERT and traditional topic embeddings from algorithms such as LDA, further passed to a linear layer and a Softmax activation. Although this framework relies on topic extraction, its main purpose is to solve semantic similarity tasks between pairs of documents and not topic extraction.

Combined Topic Model (CTM, not to be confused with Correlated Topic Models) \cite{Bianchi2021} concatenates pre-trained contextualized embeddings from SBERT and BOW embeddings in a Variational Autoencoder (VAE) \cite{Kingma2014} grounded on the idea of Autoencoded Variational Inference for Topic Model (AVITM) proposed by \cite{Srivastava2017} and its Product of Experts LDA (ProdLDA), an extension of traditional LDA to a variational neural setting. Its results demonstrate that combining both SBERT and BOW may impact the topic coherence positively.

\section{Discussion and Research Opportunities}
\label{sec:discussion}

The reviewed works did not use reinforcement, active or other machine learning paradigms for clustering or topic modeling tasks, which seem to be promising for text clustering research. For instance, the combination of neural topic modeling and reinforcement learning presented encouraging results \cite{gui-etal-2019-neural}. Furthermore, associating active learning with text clustering can also result in better performance metrics \cite{Balafar_etal_active_2020}. Besides, there are research opportunities pertaining to the automated machine learning field, particularly hyperparameter optimization approaches, such as Bayesian hyperparameter optimization \cite{terragni-etal-2021-octis}. In the context of transfer learning, there are also applications to clustering tasks \cite[p.48]{automl_book}. Moreover, when analyzing the algorithmic complexity, it would be more fair considering ranges of the inputs, length of the text vectors, or whether there are underrepresented topics (e.g., based on ground-truth labels) when choosing the most suitable algorithm for each task, instead of raw statistics of the asymptotic behavior of best, average or worst cases of the implementations. The inclusion of the likely impact of large constants on the processing time during the decision-making may improve the use of computational resources.

Since it is hard to match human intuition for topic models by automated optimization \cite{Hennig2015, Chang2009}, users may help the modeling algorithm by giving active feedback during the optimization process. This resource was not considered in the reviewed works, which seems to reflect research opportunities of interactive topic modeling and interactive outlier handling. For instance, interactive topic modeling pipelines may be employed in areas where the typical user will not have machine learning expertise to tune the model hyperparameters by hand or provide the necessary modifications directly to reflect its objectives \cite{Hu2011}. Concerning initialization, the seeding process presents itself as an alternative to more efficient approaches. The seeding process consists of methods to cherry-pick more favorable elements to begin the clustering or topic modeling to yield faster convergence, better performance, and lower computational costs. This process can be improved whenever an oracle is available (e.g., user interaction and semi-supervision). In addition, the gains provided by methods that are robust to initialization settings and the ones that arise from better seeds need to be compared. Finally, some criteria can address the specificity of the number of seeds depending on corpus size, text domain, and algorithm.

Besides, the evaluation of topic models and clustering methods that do not use a predefined number of groups can be challenging. If a method obtains more groups than the ground-truth labels (whenever a labeled corpus is available), it can mean that the modeled groups are overly specific. Nonetheless, there are challenges to be addressed, including: partitioning criteria; cluster granularity; comparison of ground-truth labels and the obtained clusters (even though there are different numbers of groups); acceptability of smaller or bigger groups in comparison to the original labels; penalization of outliers mistakenly grouped; balance of the trade-off concerning the mistakes of clustering outliers or gathering instances that belong to other groups; analysis of the correspondence between predicted and actual groups; and the fine-tuning of hyperparameters albeit the inconsistency of performance metrics.

Despite machine learning not having silver bullets, developing and testing approaches to evaluate topic models and clustering algorithms assesses structured and effective cooperation and development of text clustering research. There is a need for benchmarks that include metrics and corpora, which was also reported in \cite{harrando2021,terragni-etal-2021-octis,lisena-etal-2020-tomodapi,hoyle2021automated}. New topic models had been released in 2021. Nevertheless, due to the absence of baselines and standardized benchmarks \cite{hoyle2021automated}, defining the state-of-the-art is not a trivial task. Thus, this cumbersome issue hinders the advancement of research, leading to rework and unrealistic performance gains (mainly when there is no statistical analysis). The literature of text clustering and topic models would benefit from the definition of corpora to enable benchmarks concerning different text domains and lengths, the study of the influence of text length or its domain on the language models (LMs) and their relation with clustering quality, and the determination of standardized metrics that can be applied to older topic modeling algorithms (e.g., LDA) and to new ones (e.g., neural topic models) \cite{doogan-buntine-2021-topic}.

NMF and LDA rely on the document-term matrix to represent the text, while clustering methods such as K-means and HDBSCAN can use more general embeddings. These vectors can be sparse, and the curse of dimensionality can hamper the algorithm performance when the vector length is bigger than the corpus size. As a result, dimensionality reduction arises as an alternative to improve these algorithms. Defining the best algorithms to text data, fine-tuning their hyperparameters, and reducing their stochasticity may lead to performance gain.

Methods whose objective solely relies on anomaly detection (e.g., IF, LOF, and One-class SVM) require tuning to the corpus characteristics. Understanding the best moment to detect outliers (either before clustering or topic modeling, during the optimization process, or after the group definition), the particularities of text data, the combination of algorithms, and the tuning without depending on the users' expertise are further steps of research. \cite{Campos2016evaluationofunsupervised} and \cite{DOMINGUES2018comparativeevaluation} considered the evaluation of ensemble-based algorithms as another research opportunity. Understanding how each outlier detection algorithm behaves depending on the type of outlier is a new research direction \cite{Steinbuss2021benchmarkingunsupervised}. Active learning to select instances for labeling, transfer learning, and self-supervised learning also pose other opportunities \cite{Ruff2021unifyingreview}. Research directions include the scrutiny of the interpretability and the trustworthiness of the algorithms \cite{Ruff2021unifyingreview}. Finally, weak supervision and semi-supervised learning can improve the algorithm performance \cite{Ruff2021unifyingreview, SoenenHyperparameterTuning2021, Han2022ADBench}.

\section{Related Works}\label{sec:related-works}

\begin{figure}[t]
    \centering
    \includegraphics[scale=0.60]{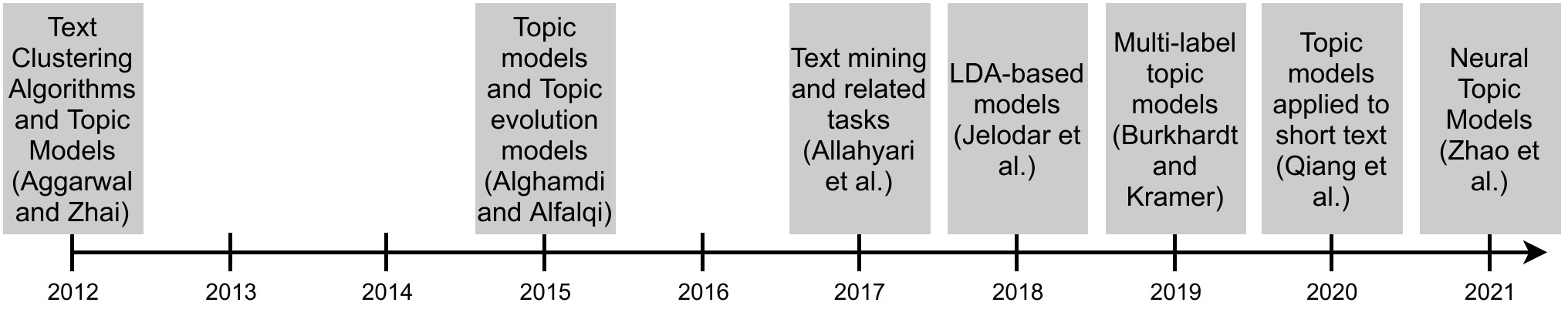}
    \caption{Timeline of Literature Reviews about Topic Modeling and Text Clustering.}
    \label{fig:timeline_surveys}
\end{figure}

We introduce this section with Figure \ref{fig:timeline_surveys}, which presents a timeline (last 10 years) with surveys about clustering and topic models. \cite{AggarwalZhai2012} introduced text clustering algorithms, including feature selection, dimension reduction, particularities of text data, distance-based algorithms (agglomerative, hierarchical, partitioning, and hybrid methods), word and phrase-based clustering (including the pattern of word frequency or phrase frequency, and the combination of word clusters and document clusters), assumptions and concepts of topic models, clustering applied to text streams and networks, and concepts of semi-supervised clustering. This work mentioned outliers in the context of temporary anomalies in clusters (e.g., online clustering), such as short-term shifts. Three years later, \cite{Alghamdi2015} presented a survey about topic models dividing them into ``topic models'' and ``topic evolution models'', which are algorithms that consider a temporal evolution of topics. Their focus was on explaining each algorithm based on its characteristics, examples, and limitations. Next, the work of \cite{allahyari2017brief} comprised a larger spectrum of text mining tasks, including text clustering (hierarchical and distance-based), text preprocessing, topic modeling (pLSA \cite{Hofmann1999} and LDA \cite{Blei2003}), information extraction (such as named entity recognition), and text classifiers. Finally, the work addressed some text mining applications in biomedicine.

In 2018, \cite{JelodarLatent2018} investigated LDA-based topic models from 2003 to 2016 that considered applications like medicine and political sciences. In addition, the LDA algorithms involved text and other data (e.g., image and audio). They also included available datasets and frameworks. In the following year, \cite{Burkhardtsurvey2019} surveyed multi-label topic models, which can also perform multi-label classification tasks. The work detailed topic models concerning sampling, online, non-parametric, and dependency methods. In addition, the paper included multi-label classification and particularities of the related topic modeling. There were presented corpora for multi-label tasks, performance metrics, and current limitations of the literature. The following review, \cite{Qiangshorttext2020}, comprised topic models applied to short texts. They presented a taxonomy for the algorithms, including Dirichlet multinomial mixture, global word co-occurrences, and self-aggregation. The work also provided an open-source tool (STTM) implemented in Java within a single interface. Finally, \cite{zhao-ijcai2021-638} shed light on neural topic models (NTMs), which emerged as competitive algorithms in the context of deep learning and NTMs challenge probabilistic topic models (e.g., LDA). The survey also provided a taxonomy for NTMs (e.g., NTMs with meta-data, NTMs for short text, sequential NTMs, NTMs with pre-trained models, NTMs based on autoregressive models, NTMs based on Generative Adversarial Nets, and NTMs based on Graph Neural Networks). Besides, the evaluation methods, limitations of NTMs, and future directions were discussed. Lastly, our work is distinct from the ones mentioned above due to its time frame and foci: reproducibility and distortion issues from initialization and outlier detection concerning text clustering and topic modeling.

\section{Declaration of interests}\label{sec:declaration_of_interests}

The authors declare that they have no known competing financial interests or personal relationships that could have appeared to influence the work reported in this paper.

\bibliographystyle{unsrt}  

\appendix

\section{Core Concept: Text Vectorization}
\label{sec:text-vectorization}

NLP problems, which include topic modeling, typically require embedding text data into numeric vectors. This enables algebraic data manipulation necessary to fit into well-known optimization frameworks. Statistical-based (\ref{subsec:statistical-based}), shallow neural-based (\ref{subsec:shallow-neural-based}), and attention neural-based (\ref{attention-Neural-Based}) methods are presented in this section.

\subsection{Statistical-Based Methods}\label{subsec:statistical-based}
Statistical embeddings represent every document based on term distributions and other statistics extracted from the corpus.

In the Bag-of-words (BOW), each document is represented by its word distribution. More formally, the document, comprising words from a vocabulary $V$, is transformed into a word frequency vector $d \in \mathbb{R}_{+}^{\lvert V\lvert}$. Major drawbacks of this embedding representation are high representational sparsity even in mildly sized corpora and loss of information regarding word order, hence ignoring word semantics based on the context in which they were originally inserted.

This simple idea can easily be further generalized to a Bag-of-$n$-grams, where each term becomes a sequence $n$ consecutive words (where word order matters) \cite{Avros2018, Li2017, Babur2016}. Increasing $n$ captures more relations between words and relieves the semantic loss from the original BOW, but at the expense of a higher dimensional representation with increased sparsity. Bag-of-$n$-grams can also be defined in character level instead of word level \cite{Deepak2018}, essentially becoming bag-of-subwords. A total of 11 (19\%) reviewed works used BOW or Bag-of-$n$-grams during their experiments.

Another popular representation is the Term Frequency-Inverse Document Frequency (TF-IDF) \cite{Luhn1957, Jones72astatistical, Salton1983IntroductionTM}, which is directly proportional to the term frequency within each document and inversely proportional on a logarithm scale to how often a term appears among all the documents in the corpora. Since the precise definition for either TF or IDF terms varies in the literature, a particular formulation for TF-IDF score is shown in Equation \ref{eq:tf-idf}.
\begin{equation}
    \label{eq:tf-idf}
    \tfidf(w, d; f, D, p) =
    \underbrace{
        \frac{f_{(w,d)}}{
            \left(
            \sum_{z \in d}^{} f_{(z, d)}^{\;p}
            \right)^\frac{1}{p}
        }
    }_{\tf(w,\ d;\ f, p)}
    \cdot
    \underbrace{
        \vphantom{
        \frac{1}{
            \left(
            \sum f_{(z, d)}^{\;p}
            \right)^\frac{1}{p}
        }
        }
        \left[
        1 -
        \log\left(\frac{n(w; D) + 1}{\lvert D\lvert + 1}\right)
        \right]
    }_{\idf(w;\ D)}
\end{equation}
In this equation, $D$ is the set of all documents in the corpus, $w$ is a particular term in vocabulary $V$, $d$ is a particular document in $D$, $f \in \mathbb{R}_{+}^{\lvert V \lvert \times \lvert D \lvert}$ is the frequency matrix for every term-document pair, $p > 0$ defines the normalization type to the term frequency component (typically $p=1$ or $p=2$ for $\ell_1$ and $\ell_2$ normalization respectively), and $n : V \rightarrow \mathbb{N}$ maps every word $w$ to the count of documents where it appears at least once, as shown in Equation \ref{eq:word-document-frequency}, where $\indicatorop$ is an indicator function.
\begin{equation}
    \label{eq:word-document-frequency}
    n(w; D) = \sum_{d \in D}\indicator{w \in d}
\end{equation}
A commonly used implementation of such embedding is available in the scikit-learn library \cite{scikit-learn, sklearn_api}, which closely follows the definition in Equation \ref{eq:tf-idf} given its default parameters, with the single difference being a final document score normalization, as shown in Equation \ref{eq:tf-idf-scikit} with $p = 2$.
\begin{equation}
    \label{eq:tf-idf-scikit}
    \tfidf_\text{scikit}(w, d; f, D, p) =
    \frac{
        \tfidf(w, d; f, D, p)
    }{
        \left(
            \sum_{z\in d} \tfidf(z, d; f, D, p)^{p}
        \right)^\frac{1}{p}
    }
\end{equation}
The rationale behind this definition (and many other similar ones in the literature) is that terms that frequently occur in a document $d$ and are rarer in other documents are good representatives of $d$. A total of 17 (30\%) reviewed works used TF-IDF as the embedding representation, making it the most popular one among the reviewed works.

\subsection{Shallow Neural-Based Methods}\label{subsec:shallow-neural-based}
Neural-based methods benefit from using neural networks to learn the representation of text data as dense vectors, also known as distributed representations \cite{Bengio2003}, which capture semantic and syntactic properties and relationships between objects from similarity scores. These types of embeddings are also often pre-trained in large, general-purpose corpora and made publicly available. Pre-trained embeddings are widely used in topic modeling. Another advantage of neural-based embeddings is that their dimensionality is generally much smaller than traditional BOW or TF-IDF embeddings, which helps with the dimensionality issues in data clustering and factorization and also reduces representational sparsity.

The self-supervised framework word2vec \cite{mikolov2013distributed} and \cite{mikolov2013efficient} uses shallow neural networks to create word embeddings. It is implemented by two distinct algorithms: Continuous Bag-of-Words (CBOW) and Skip-Grams. Let $C_{t} = \{w_{(t-i)} \lvert -n \leq i \leq n \land i \neq 0\}$ be the context window to the $t$-th word $w_{t}$. CBOW works by predicting $w_{t}$ from $C_{t}$ as input, while the Skip-Grams predict $C_{t}$ from $w_{t}$ as input. This last approach is often implemented as multiple independent binary classification problems, where the model learns to decide whether words are part of $C_{t}$ or part of a randomly sampled subset of words $N_{t}$ unrelated to $w_{t}$. This approach is called Skip-Grams with Negative Sampling \cite{mikolov2013distributed}, and its objective function is shown in Equation \ref{eq:word2vec-skipgram}. It is mathematically related to the Pointwise Mutual Information matrix \cite{Levy2014}. From the reviewed works, 9 (16\%) used word2vec as the embedding method during the experiments.
\begin{equation}
    \label{eq:word2vec-skipgram}
    J(w; D) =
    \sum_{w_t \in D}
    \left[
        \sum_{w_j \in C_{t}} \log(1 + \exp(-w_{t}^{T} w_{j}))
        +
        \sum_{w_j \in N_{t}} \log(1 + \exp(w_{t}^{T} w_{j}))
    \right]
\end{equation}
By observing that the co-occurrence matrix of words in $V$ is very sparse, GloVe (Global Vectors) \cite{pennington2014glove} incorporates co-occurrence statistics directly into its objective function, shown in Equation \ref{eq:glove}, where $X_{(i, j)}$ is the count of co-occurrences for words $w_{i}$ and $w_{j}$ within the same context windows. While the original work claims this modification makes GloVe computationally cheaper to train and outperform word2vec in certain tasks, conclusions from \cite{Levy2015} are the opposite when both algorithms receive proper hyperparameter tuning or are used in huge corpora. From the reviewed works, 5 (9\%) used GloVe as the embedding method in its experiments.
\begin{equation}
    \label{eq:glove}
    J(w, b; X) =
    \sum_{i,j=1}^{\lvert V \lvert}
    \min
    \Bigl(
        1, 
        \Bigl(
            \frac{X_{(i,j)}}{100}
        \Bigr)^{0.75}
    \Bigr)
    \left(
        w_{i}^{T} w_{j} + b_{i} + b{j} - \log X_{(i,j)}
    \right)^{2}
\end{equation}
So far, the previous methods assigned a single vector to each word, which is difficult to learn good embedding for rare words, cannot generalize to out-of-vocabulary words, and do not leverage shared morphemes across words (e.g., affixes). Thus, fastText \cite{bojanowski2017enriching, joulin2016bag, mikolov2018advances} addresses these issues by learning embeddings for \textit{subwords} defined as $n$-grams of individual characters and then representing a whole word by summing up the embeddings of every piece that composes it. fastText is trained using the word2vec (Skip-Grams) framework. Only 2 (4\%) works used fastText as their embedding method during the experiments. All of the aforementioned embedding methods do not address polysemy, where a single word can have very distinct meanings depending on its context.

Aiming to embed the entire documents instead of single tokens, doc2vec \cite{le2014distributed} was developed based on the word2vec idea. This framework can also be implemented with two distinct algorithms: Distributed Memory Model of Paragraph Vectors (PV-DM) and Distributed Bag-of-Words version of Paragraph Vector (PV-DBOW), which are analogous to CBOW and Skip-Grams, respectively. The main difference between this method and its word counterpart is that doc2vec also trains unique embeddings to every document, called \textit{paragraph vectors}, alongside the usual word embeddings which are shared across all documents. All learned parameters are fixed after training, but paragraph vectors for unseen documents must be learned using gradient descent, which can lead to relatively high inference costs. From the reviewed works, 3 (5\%) used doc2vec in their experiments.

\subsection{Attention Neural-Based Methods}\label{attention-Neural-Based}
In the field of NLP, the concept of attention in machine learning emerged in \cite{Bahdanau2016} for sequence-to-sequence (seq2seq) alignment models, having captured much research interest, and therefore was developed in many nuances and now is applied to multiple tasks \cite{Chaudhari2021, Correia2021}. One of the simplest and most popular variants is the scaled dot-product (or QKV) attention as shown in Equation \ref{eq:scaled-qkv-attention}, correlating every pair of tokens from the input sequences. In that setting, $Q \in \mathbb{R}^{N \times E}$ and $K, V \in \mathbb{R}^{M \times E}$, where $N$ is the length of the first input sequence, $M$ is the length of the second input sequence, and $E$ is the embedding dimension for both. The Softmax activation is taken independently for each row of its argument, thus building a probabilistic distribution for each input token from the first sentence by taking into account every token from the second sentence.
\begin{equation}
    \label{eq:scaled-qkv-attention}
    \attention(Q, K, V) = \softmax\left(\frac{Q K^{T}}{\sqrt{E}}\right) V
\end{equation}
Heavily based on the machine attention concept, Transformers \cite{Vaswani2017} were introduced, rapidly turning into the state-of-the-art algorithm for seq2seq processing. Their encoder-decoder neural network architecture comprises identical stacked blocks with built-in attention mechanisms and non-linear activation. With them, the concept of multi-headed Attention was also introduced, shown in Equation \ref{eq:multiheaded-attention} where $h$ is the number of attention heads, $W_{i}^{Q}, W_{i}^{K}, W_{i}^{V} \in \mathbb{R}^{E \times (E/h)}$ for $i=1, \ldots, h$, and $W_O \in \mathbb{R}^{E \times E}$. Each head learns to match distinct features from the input distribution and, later, linearly combined by $W_O$ to form each transformer block output.
\small
\begin{equation}
    {\mattention(Q, K, V; W) =
    \left[
    \attentionp{1}
    ,\,
    \dots
    ,\,
    \attentionp{h}
    \right]
    \cdot
    W_{O}}
    \label{eq:multiheaded-attention}
\end{equation}\normalsize
Transformers relieved the problem of processing large chunks of text of previous seq2seq models since they process every token in parallel. However, that characteristic lacks word order information. Thus, initial transformer embeddings are combined with positional embeddings, which uniquely identify each token position. Finally, due to the chosen attention mechanism, every token embedding is built by using global information from the entire input. By combining all described characteristics, transformers are able to create embeddings that actually take into account how and where tokens are inserted into the sentences, which is why they are called \textit{contextual embeddings}.

Equation \ref{eq:contextual-embeddings} summarizes all described embedding characteristics, where $f$ is a function that combines both token and positional embeddings and typically consists of an addition operation. One advantage of contextual embeddings is that they can better represent polysemous words, characteristic that previously presented embedding algorithms fail to address.
\begin{equation}
    \label{eq:contextual-embeddings}
    e_\text{contextual}(w, d; f) = f(e_\text{token}(w, d), e_\text{positional}(\text{index}(w, d)))
\end{equation}
The ubiquitous Bidirectional Encoder Representations from
Transformers (BERT) \cite{Devlin2019} is a encoder-only transformer pretrained on a general-purpose English corpus (BooksCorpus \cite{Zhu2015} and the English Wikipedia) simultaneously in two self-supervised surrogate tasks, Masked Language Modeling (MLM) and Next Sentence Prediction (NSP), and made publicly available as 12-layers (base size) or 24-layers (large size) models. Currently, it is the most popular attention-based architecture for transfer learning in the field of natural language processing, being fine-tuned to many downstream tasks. BERT assumes the input sequence is tokenized as sub-word representations from WordPiece algorithm \cite{Wu2016}, thus creating embeddings for such tokens and not for individual words.

Let $T_\text{BERT}(\ \cdot\ ; \theta)$ be a BERT model with trainable weights $\theta$. Its output can be used as token-wise embeddings for document $d$, $T_\text{BERT}(d; \theta) = [e_0,\ e_1, \ldots, e_{\lvert d\lvert}] \in \mathbb{R}^{E \times (1 + \lvert d\lvert)}$, where $\lvert d\lvert$ is the sub-word count for document $d$ and $e_0$ consists of the embedding corresponding to the NSP output. For a fixed-sized document representation $e \in \mathbb{R}^{E \times 1}$ one can, for instance, take the element-wise average of sub-word embeddings (discarding the NSP embedding $e_0$), also known as ``average pooling operation'', as shown in Equation \ref{eq:BERT-embeddings}. Another way is to adopt $e_0$ as the document embedding.
\begin{equation}
    \label{eq:BERT-embeddings} 
    e = \frac{1}{\lvert d\lvert}\sum_{i=1}^{\lvert d\lvert} e_{i}
\end{equation}
Although one can use $T_\text{BERT}(d; \theta)$ directly for downstream tasks, the output from transformer encoder final blocks are often specialized to the pretraining task output space and may perform poorly on unrelated downstream tasks without any fine-tuning \cite{Liu2019}. The output from one or a linear combination of multiple intermediary blocks tends to encode more general language features \cite{Merchant2020, Liu2019}, and thus are more appropriate to unrelated downstream tasks. For similarity-related tasks, very typical in the context of topic modeling, original BERT embeddings are remarkably unsuccessful, even worse than averaged GloVe embeddings \cite{Reimers2019}.

Using transfer learning from the original pretrained BERT model, Sentence-BERT (SBERT) \cite{Reimers2019} proposes a training procedure to create models specialized in capturing similarity relationships between sentences (or small documents). This algorithm optimizes a siamese network architecture in a surrogate supervised classification task of Natural Language Inference (NLI), with 1 million sentence pairs manually annotated as \textit{contradiction}, \textit{entailment} or \textit{neutral} (Stanford Natural Language Inference \cite{Bowman2015} and Multi-Genre NLI Corpus \cite{Williams2018}). The objective function adopted is the cross-entropy loss as shown in Equation \ref{eq:SBERT-objective}, where $T_\text{SBERT}(\,\cdot\,; \theta)$ is a SBERT model with trainable parameters $\theta$, $(d_{(i, a)}, d_{(i, b)})$ is the $i$-th document pair from a paired (or parallel) corpora $D$, $y \in \{0, 1\}^{\lvert D\lvert \times 3}$ are the one-hot encoded labels from NLI classes, and $\phi \in \mathbb{R}^{3 \times 3E}$ are learnable parameters used to mix and project the document pair embeddings onto the NLI task output space and discarded after training. The logarithm and absolute value functions are applied element-wise.
\begin{equation}
    \label{eq:SBERT-objective}
    J(\theta, \phi; D, y, T_\text{SBERT}) =
    - \displaystyle\sum_{i=1}^{\lvert D \lvert}
    y_{(i)}
    \cdot
    \log
    \Bigl(
    \underbrace{
        \hat{y} (
            \overbrace{
                T_\text{SBERT}(d_{(i,a)}; \theta)
            }^\text{$e_a$: $d_a$ embedding}
            ,
            \overbrace{
                T_\text{SBERT}(d_{(i, b)}; \theta)
            }^\text{$e_b$: $d_b$ embedding}
        ; \phi)
    }_{
        \hat{y}(e_a,\ e_b;\ \phi)\ =\ \softmax
        \left(
        \phi
        \cdot
        [
        e_a,\ e_b,\ \lvert e_a\ -\ e_b \lvert
        ]
        \right)
    }
    \Bigr)
\end{equation}
To transform every piece of text $d$ into a fixed-length embedding $e_d$, an average pooling layer is attached to the end of SBERT (see Equation \ref{eq:BERT-embeddings}). Unlike the original BERT embeddings, SBERT embeddings are appropriate for similarity-based tasks such as document clustering and outlier detection. Section \ref{sec:end-to-end-topic-models} reviewed end-to-end topic models with pre-trained transformer encoders.

\section{Fundamentals of Factorization Algorithms}
\label{sec:results-factorization-methods}

Matrix factorization can be employed as a direct topic modeling algorithm. In addition, it can be used as a dimensionality reduction algorithm in a data processing pipeline to overcome the curse of dimensionality, which is often present in text embedding methods. This appendix presents Latent Semantic Analysis (LSA) in \ref{subsec:lsa}, Non-negative Matrix Factorization (NMF) in  \ref{subsec:nmf}, and Latent Dirichlet Allocation (LDA) in \ref{subsec:lda}.

\subsection{Latent Semantic Analysis (LSA)}\label{subsec:lsa}
Also known as Latent Semantic Indexing (LSI) \cite{Deerwester1990}, LSA is a simple factorization algorithm where the embedded corpus $D$, traditionally by BOW or TF-IDF (Equation \ref{eq:tf-idf}), is factorized by Singular Value Decomposition (SVD) as shown in Equation \ref{eq:svd}, where $r = \text{rank}(D)$ denotes the number of independent features in the embedding of $D$, $U \in \mathbb{R}^{n \times r}$ denotes an orthonormal document-topic matrix, $\Sigma \in \mathbb{R}_{+}^{r \times r}$ is a non-negative diagonal matrix and by convention $\sigma_{1} \geq \dots \geq \sigma_{r}$, and $V \in \mathbb{R}^{m \times r}$ is an orthonormal feature-topic matrix.
\begin{equation}
    \label{eq:svd}
    \overset{D_{(n, m)}}{
    \underbrace{
        \left[
            \begin{array}{ c }
                \horzbar\ d_{1}\ \horzbar \\
                          \vdots          \\
                \horzbar\ d_{n}\ \horzbar \\
            \end{array}
        \right]
    }_{\substack{\text{Document-feature matrix:} \\ \text{Original document embeddings}}}
    }
    \hspace{0.25cm}
    =
    \overset{U_{(n, r)}}{
    \underbrace{
        \left[
            \begin{array}{ c }
                \smallhorzbar\ d_{1}^{'}\ \smallhorzbar \\
                          \vdots               \\
                \smallhorzbar\ d_{n}^{'}\ \smallhorzbar \\
            \end{array}
        \right]
    }_{\substack{\text{Document-topic matrix:} \\ \text{Documents embedded} \\ \text{in topic subspace}}}
    }
    \cdot
    \hspace{0.25cm}
    \overset{\Sigma_{(r, r)}}{
    \underbrace{
        \vphantom{
        \left[
            \begin{array}{ c }
                \smallhorzbar\ d_{1}^{'}\ \smallhorzbar \\
                          \vdots               \\
                \smallhorzbar\ d_{n}^{'}\ \smallhorzbar \\
            \end{array}
        \right]
        }
        \left[
            \begin{array}{c c c}
                \sigma_{1} &        &            \\
                           & \ddots &            \\
                           &        & \sigma_{r} \\
            \end{array}
        \right]
    }_{\substack{\text{Topic relevance matrix:} \\ \text{Describes how important} \\ \text{every topic is}}}
    }
    \hspace{0.2cm}
    \cdot
    \overset{V_{(r, m)}^{T}}{
    \underbrace{
        \vphantom{
        \left[
            \begin{array}{ c }
                \smallhorzbar\ d_{1}^{'}\ \smallhorzbar \\
                          \vdots               \\
                \smallhorzbar\ d_{n}^{'}\ \smallhorzbar \\
            \end{array}
        \right]
        }
        \left[
            \begin{array}{c c c}
                \smallvertbar &        & \smallvertbar \\
                v_{1}    & \cdots & v_{m}    \\
                \smallvertbar &        & \smallvertbar \\
            \end{array}
        \right]
    }_{\substack{\text{Topic-feature matrix:} \\ \text{maps documents} \\ \text{to topic subspace}}}
    }
\end{equation}
The SVD decomposition can always be performed to any matrix $D$ and is unique up to permutations and signals. We can choose any number of topics $k$ as long as $1 \leq k \leq r$ holds. If $k < r$, then Equation \ref{eq:svd} becomes a rank-$k$ approximation to $D$ with the least Frobenius-norm error \cite{Eckart1936}.

From a geometrical standpoint, the columns in $U$ compose an orthonormal vector basis of the embedding space of $D$ (when $k = r$). Thus, $U$ can represent documents in a potentially compressed embedding that spans the topic subspace, removing all the redundancy present in the original embedding. The matrix $V$ describes the linear transformation that maps $D$ into the topic subspace, $D \xrightarrow{V} U$, and $\Sigma$ keeps $U$ orthonormal, absorbing the variability inherently to each feature in the original embedding so that topics with a high corresponding $\sigma_{i}$ encode more corpus information and thus can describe documents better. When $k < r$, we approximate this relationship by only keeping the top-$k$ most relevant topics.

This formulation can be generalized by using $\Sigma^p$, for some sharpness hyperparameter $p \in \mathbb{R}$, controlling the relative importance of principal components (topics) \cite{Caron0000, Levy2015}. In fact, representing $D$ in the topic space as $U\Sigma$ or just $U$ is a matter of setting $p = 1$ or $p = 0$, respectively. Larger $p$ emphasizes more discriminative topics, while $p \rightarrow 0$ dilutes the relevance across topics. This hyperparameter requires tuning, but $p \rightarrow 0$ performs better in word similarity tasks \cite{Levy2015}.

The Probabilistic LSA/LSI (pLSA/pLSI) \cite{Hofmann1999} is a further development from the LSA, posing the problem as to describe corpus $D$ with a probabilistic algorithm. Its objective function is shown in Equation \ref{eq:plsa} as the negative log-likelihood of $D$, where $f \in \mathbb{R}^{\lvert V\lvert \times \lvert D\lvert}$ denotes the term-document matrix of corpora $D$ with vocabulary $V$, and $z$ is a set of unknown topics such that $\lvert z\lvert$ (the number of topics) is given and fixed. The trainable model parameters are $\theta$ and $\phi$.

\begin{equation}
    \label{eq:plsa}
    J(\theta, \phi; D, f) =
    -\log\pparam{D}{\theta, \phi, f} =
    -\sum_{d \in D}^{}
    \sum_{w \in d}^{}
    f_{(w, d)}
    \log
    \left(
    \sum_{z}^{}
    p(w \lvert z;\phi_{(z,w)})
    p(z \lvert d;\theta_{(z,d)})
    \right)
\end{equation}
This model has a few clear drawbacks. Firstly, the model parameter count $\lvert \theta \lvert$ grows linearly with corpus size $\lvert D\lvert$, $\lvert \theta \lvert$ $\in \mathcal{O}(\lvert D\lvert)$, making the algorithm infeasible for large corpora and raising concerns of parameter overfitting to the data \cite{Blei2003}. The reviewed work \cite{Veselova2020} proposes a regularization strategy for pLSA. Moreover, unseen documents outside $D$ are not modeled by pLSA. From the reviewed works, 3 (5\%) used LSA/LSI in their experiments, and 2 (4\%) used pLSA/pLSI.

\subsection{Non-negative Matrix Factorization (NMF)}\label{subsec:nmf}
The objective of NMF \cite{Paatero1994, Daniel1999} is approximating a non-negative matrix \nnmatwithdims{D}{n}{m} by two lower rank non-negative matrices \nnmatwithdims{W}{n}{k} and \nnmatwithdims{H}{k}{m}, where $k < m$ is a hyperparameter that controls the trade-off between representation precision and computational cost. In the context of topic modeling, $D$ is the embedded corpus by non-negative features, and $k$ is the predefined number of latent topics in $D$. NMF is not unique in general and, therefore, is affected by the initial conditions \cite{Febrissy2020, Hassani2021}. It is typically done by minimizing the constrained objective function expressed by Equation \ref{eq:nmf}, where $\Vert \cdot \Vert_{F}$ denotes the Frobenius matrix norm. The Kullback-Leibler Divergence \cite{Kullback1951} may also be used \cite{Febrissy2020}.
\begin{align}
    \begin{split}
        \label{eq:nmf}
        \min\limits_{W,H} J(W, H; D) &
        = \min\limits_{W,H}
        \diffnorm{D}{WH}_{F}^{2}
    \end{split} \\
    \text{subject to}
        & \quad W, H \geq 0 \notag
\end{align}
There are several algorithms to find good solution candidates \cite{Kim2014}, such as projected gradient descent \cite{Lin2007, Chih2007}, coordinate descent \cite{Hsieh2011}, or scaled gradient descent as shown in Equation \ref{eq:scaled-gradient-descent}, where $\theta \in \mathbb{R}^{n \times m}$ are trainable parameters, $\eta \in \mathbb{R}_{+}^{n \times m}$ are element-wise learning rates and $\odot$ denotes matrix element-wise (or Hadamard) product.
\begin{equation}
    \label{eq:scaled-gradient-descent}
    \theta^{(t + 1)} =
    \theta^{(t)} - \eta \odot \nabla_{\theta^{(t)}}J(\theta^{(t)})
\end{equation}
One scaled gradient descent particular formulation to find candidate solutions for NMF is known as the Multiplicative Update (MU) \cite{NIPS2000_f9d11525}, where particular element-wise learning rates $\eta$ are chosen and plugged into Equation \ref{eq:scaled-gradient-descent}, which guarantee that both $W$ and $H$ stay non-negative after every update. The MU rules are shown in Equation \ref{eq:multiplicative-update}, where divisions are also element-wise. Although simple to present and implement, this approach has a slow convergence rate, and more sophisticated algorithms should be preferred \cite{Gonzalez2005, Kim2014}.
\begin{equation}
    \label{eq:multiplicative-update}
    W_{\text{new}} = W \odot \frac{D H^{T}}{W H H^{T}}
    \qquad\qquad
    H_{\text{new}} = H \odot \frac{W^{T} D}{W^{T} W H}
\end{equation}
From the perspective of topic modeling, NMF learns how to build compact representations of each document (matrix $W$) as specific mixtures (given by matrix $H$) of a predetermined number of topics ($k$), as depicted in Equation \ref{eq:nmf-topic-modeling}.
\begin{equation}
    \label{eq:nmf-topic-modeling}
    \overset{D_{(n, m)}}{
    \underbrace{
        \left[
            \begin{array}{ c }
                \horzbar\ d_{1}\ \horzbar \\
                          \vdots          \\
                \horzbar\ d_{n}\ \horzbar \\
            \end{array}
        \right]
    }_{\substack{\text{Document-feature matrix:} \\ \text{Original document embeddings}}}
    }
    \hspace{0.25cm}
    =
    \overset{W_{(n, k)}}{
    \underbrace{
        \left[
            \begin{array}{ c }
                \smallhorzbar\ d_{1}^{'}\ \smallhorzbar \\
                          \vdots               \\
                \smallhorzbar\ d_{n}^{'}\ \smallhorzbar \\
            \end{array}
        \right]
    }_{\substack{\text{Document-topic matrix:} \\ \text{Documents embedded} \\ \text{in topic subspace}}}
    }
    \cdot
    \hspace{0.25cm}
    \overset{H_{(k, m)}}{
    \underbrace{
        \vphantom{
        \left[
            \begin{array}{ c }
                \smallhorzbar\ d_{1}^{'}\ \smallhorzbar \\
                          \vdots               \\
                \smallhorzbar\ d_{n}^{'}\ \smallhorzbar \\
            \end{array}
        \right]
        }
        \left[
            \begin{array}{c c c}
                \smallvertbar &        & \smallvertbar \\
                h_{1}    & \cdots & h_{m}    \\
                \smallvertbar &        & \smallvertbar \\
            \end{array}
        \right]
    }_{\substack{\text{Topic-feature matrix:} \\ \text{$m$ parallel learned} \\ \text{topic mixtures}}}
    }
\end{equation}
The interpretation of NMF results may be non-trivial to topic modeling since NMF alone does not develop the concrete meaning for each topic, demanding further analysis of a data domain expert. Moreover, $k$ could be chosen inappropriately by overestimating or underestimating the ``true'' topic number, hence reducing topic coherence.

\subsection{Latent Dirichlet Allocation (LDA)}\label{subsec:lda}
\begin{figure}
    \centering
    \includegraphics[scale=0.6]{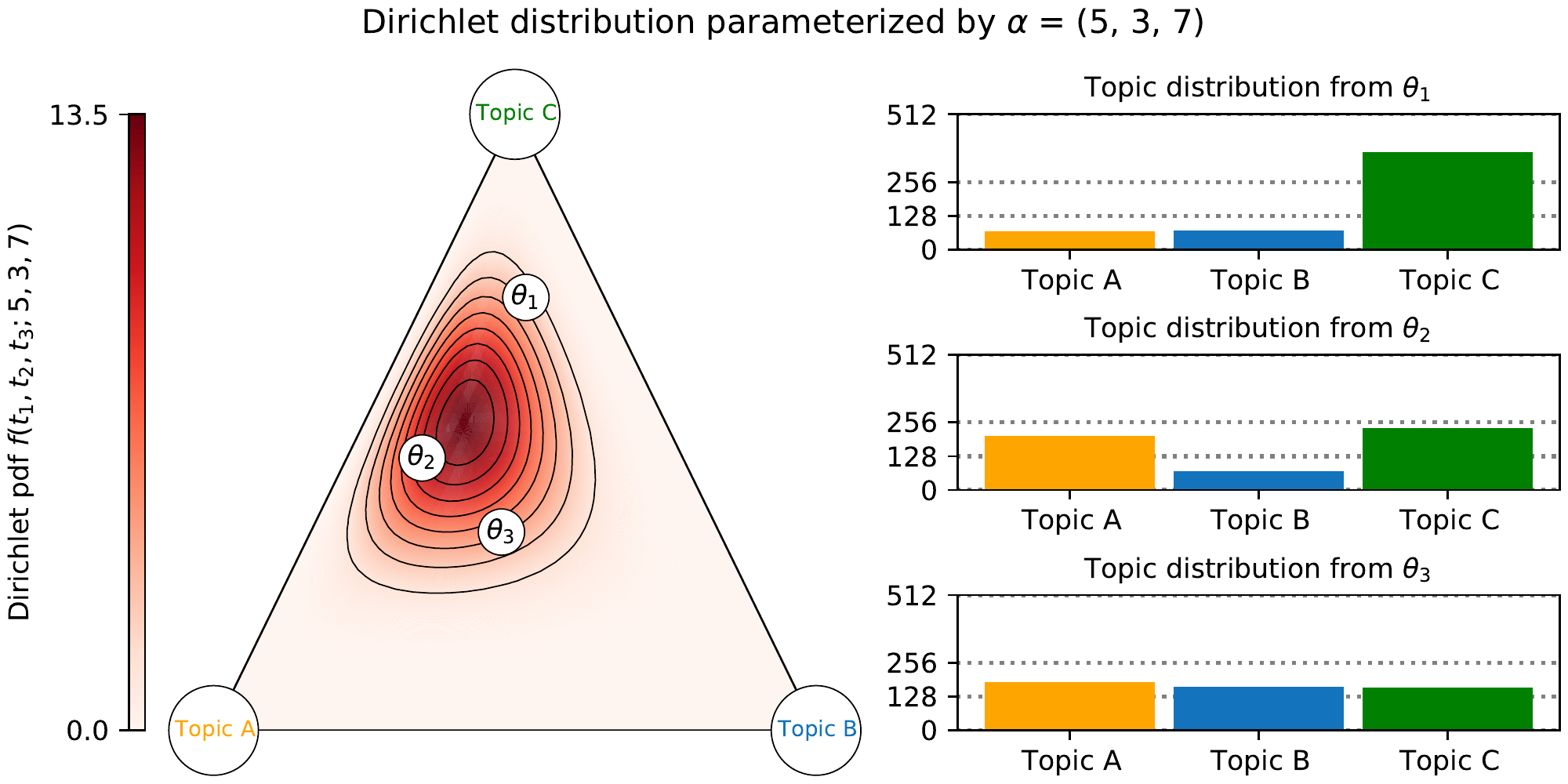}
    \caption{Example of a Dirichlet distribution for $\alpha = (5, 3, 7)$. Each topic is represented by a corner of a 2-simplex. Three multinomial distribution parameters $\theta_{i}$ were drawn randomly from this Dirichlet distribution and plotted in the contour plot to the left. Then, for each multinomial distribution parameterized by $\theta_{i}$, 512 samples were drawn randomly and their frequencies plotted as the bar plots to the right.}
    \label{fig:dirichlet-example}
\end{figure}

LDA \cite{Blei2003} describes a probabilistic generative algorithm by a hierarchical bayesian model, where documents are modeled as topic mixtures, and topics are modeled as word mixtures. One common description of LDA is as follows:

\begin{enumerate}
    \item Let $D$ be text corpora with a vocabulary $V$. Choose and fix the number of topics $k$ and the desired generated document length $N$;
    \item Choose and fix the hyperparameters $\alpha \in \mathbb{R}^{k}$ and $\beta \in \mathbb{R}^{V}$. Typically both vectors are chosen to have all entries with the same value (which leads to symmetric Dirichlet distributions and thus balance between topics/words) and strictly less than 1 (which leads to sparse Dirichlet distributions, corresponding to more homogeneous generated documents);
    \item Sample $\theta_{i} \sim \dirichletp{\alpha}$, where $i=1, \ldots, \lvert D \lvert$;
    \item Sample $\phi_{j} \sim \dirichletp{\beta}$, where $j=1, \dots, k$;
    \item Repeat for every $i=1, \dots, \lvert D \lvert$ and $j=1, \dots, N$:
    \begin{enumerate}
        \item Sample a topic $z_{(i,j)} \sim \multinomp{\theta_{i}}$;
        \item Given the chosen topic $z_{(i,j)}$, sample a word $w_{(i,j)} \sim \multinomp{\phi_{z_{(i,j)}}}$.
    \end{enumerate}
\end{enumerate}
LDA makes the ``bag-of-words'' (or ``exchangeability of words'') assumption, and, therefore, word order is not considered. It is a further generalization of pLSA/pLSI (see the objective function in Equation \ref{eq:plsa}), describing a document-generation process even to unseen documents, unlike its predecessor, and also does not grow its parameter count linearly with corpus size. Its objective function is formally described by Equation \ref{eq:lda} as the corpus negative log-likelihood, although it is intractable to compute. Thus, various algorithms were employed to approximate a solution, such as variational expectation-maximization \cite{Blei2003}, Gibbs sampling \cite{Griffiths5228}, and online variational bayes \cite{Hoffman2010} that scales for large corpora by using mini-batch gradient descent.
\begin{equation}
    \centering
    \label{eq:lda}
    J(\alpha, \beta; D) =
    -\log\pparam{D}{\alpha, \beta} =
    -\sum_{i=1}^{\lvert D \lvert}
    \log
    \int
    \pparam{\theta_{i}}{\alpha}
    \left(
        \prod_{j=1}^{N}
        \sum_{z_{(i,j)}}^{}
        p(z_{(i,j)};\theta_{i})
        p(w_{(i,j)} \lvert z_{(i,j)};\beta)
    \right)
    d\theta_i
\end{equation}
The behavior of a $\dirichletp{\alpha}$ distribution for topic sampling among 3 possible options is illustrated in Figure \ref{fig:dirichlet-example}. From that Dirichlet distribution (represented by the contour lines within the 2-simplex), $\theta_{i}$ parameters for multinomial distributions are drawn, which in turn define a particular distribution of topics. This process is analogous for word sampling, by using the $\dirichletp{\beta}$ distribution.

One LDA limitation is that it implicitly assumes independence between topics, which may not be reasonable in some applications. To fix this issue, slightly modified algorithms can be employed in place of LDA such as Correlated Topic Models (CTM) \cite{Lafferty2006}, where the topic Dirichlet distribution is replaced by a logistic Gaussian distribution (the topic multinomial distribution is parameterized by softmaxed samples from a multivariate Gaussian distribution), hence also modeling the correlation between topics in the Gaussian covariance matrix. Note that this extra representational capability does not necessarily correlate positively with human intuition for coherent topics \cite{Chang2009}. Other LDA variants are reviewed in \cite{JelodarLatent2018}.

Similar to the NMF approach, LDA also takes for granted the number of topics $k$, and the interpretation of its results may be non-trivial since the exact meaning for each topic needs to be defined by a data domain expert. From the reviewed works, 6 (11\%) used LDA as a topic modeling algorithm, and 5 (9\%) used LDA to produce intermediary representations in more sophisticated pipelines.

\section{Core Concepts of Clustering Algorithms}
\label{sec:results-clustering-methods}

Topic modeling often relies on clustering algorithms to group similar document embeddings into the same clusters, which are interpreted as \textit{topics}. Despite the fact that the meaning for a ``cluster'' is not always clear \cite{Castro2002, Hennig2015}, current unsupervised topic modeling has shown to be able to handle natural language text data that frequently matches some of our human intuition about how the data should be grouped (and how it should not). This section presents popular clustering algorithms among the reviewed works, summarized in Table \ref{table:clustering-algs}.

We also included each algorithm asymptotic time complexity, an important factor to consider while designing computationally scalable topic modeling pipeline. Our asymptotic notation follows set-theoretic definitions from  \cite{Cormen2009}, hence $f(n) \in \mathcal{O}(n)$ denoting the conventional $f(n) = \mathcal{O}(n)$ for a function $f$, and $\mathcal{O}(f(n)) \subseteq \mathcal{O}(g(n))$ denoting $\mathcal{O}(f(n)) \leq \mathcal{O}(g(n))$ for some $f(n) \in \mathcal{O}(g(n))$. The embedding dimension is omitted from the asymptotic analysis for simplicity, but it is worth noting that compact representations also improves running time, as distance or similarity computations are necessary to group instances.

\begin{table}
    \caption{Taxonomy of reviewed clustering algorithms in this present work.}
    \begin{tabular}{l|l|l|}
        \cline{2-3}
        & Flat clusters & Hierarchical clusters \\
        \hline
        \multicolumn{1}{|l|}{\multirow{6}{*}{Representative-based}} & K-means & HAC (single-linkage) \\
        \multicolumn{1}{|l|}{} & Spherical K-means & HAC (complete-linkage) \\
        \multicolumn{1}{|l|}{} & Fuzzy K-means & HAC (group average-linkage) \\
        \multicolumn{1}{|l|}{} & K-medoids (PAM/FasterPAM) & HAC (Ward's method) \\
        \multicolumn{1}{|l|}{} & CLARA/FasterCLARA & \\
        \multicolumn{1}{|l|}{} & CLARANS/FasterCLARANS & \\
        \hline
        \multicolumn{1}{|l|}{Density-based} & DBSCAN & HDBSCAN \\
        \hline
    \end{tabular}
    \label{table:clustering-algs}
\end{table}

This appendix is divided as follows: representative-based clustering (\ref{subsec:representative-based}), hierarchical clustering (\ref{subsec:hierarchical-clustering}), density-based clustering (\ref{subsec:density-based}), and the connection with factorization algorithms (\ref{subsec:connection}).

\subsection{Representative-Based Clustering}\label{subsec:representative-based}
Algorithms within this class compute similarity scores (or distances) between instances to build a representative object for each cluster. Equivalently, every document is represented by a similar representative object or by a weighted average of representatives proportional to the degree of similarity.

K-means \cite{Bock2007} is the most popular clustering algorithm among the reviewed works (17 (30\%) total works), used as part of the topic modeling pipeline or a baseline algorithm for more sophisticated ones. It is a partitioning algorithm, hence attributing each document to precisely one cluster. Two key assumptions of this algorithm are that the maximum number of clusters or topics $k$ is known and fixed in advance (which is not necessarily a reasonable assumption), and each topic follows a distribution with equal variance in every dimension. Geometrically, it means that topics should, presumably, be modeled as hyperspherical document clouds with the same radii.

Formally, the k-means algorithm minimizes the objective function shown in Equation \ref{eq:k-means}, the sum of squared Euclidean distances between every embedded document $d$ and its cluster center of mass $\mu \in C$, known as \textit{centroid} or \textit{cluster prototype}, with $C \in \mathbb{R}^{k \times \lvert d\lvert}$. Note that this objective function is not equivalent to minimizing the Euclidean distances between the documents and their cluster centers, which consists of a different and harder problem known as the Fermat-Weber Problem \cite{Fritz2010}. It is, however, equivalent to minimizing cluster weighted intra-variances.
\begin{equation}
    \label{eq:k-means}
    J(C; D) =
    \sum_{d \in D}
    \min\limits_{\mu \in C}
    \diffnorm{d}{\mu}_{2}^{2}
\end{equation}
Since finding the global minimum of this objective function is NP-hard \cite{Aloise2009, Mahajan2009}, heuristics are employed to find local minima, often posed as coordinate ascent algorithms such as the Lloyd's Algorithm \cite{Lloyd1982} or more sophisticated and faster ones \cite{Hamerly2017}. The Lloyd's Algorithm has an exponential worst-case time complexity of $2^{\Omega(\lvert D \lvert)}$ against adversarial examples \cite{Vattani2008}, but, in practice, it tends to run in linear average time complexity of $\mathcal{O}(Nk \lvert D \lvert)$, where $N$ is the maximum number of optimization steps.

A soft-clustering alternative for K-means is Fuzzy K-means \cite{Bezdek1981, Bezdek1984}, also known as Fuzzy C-means. In this algorithm, each document is inserted into every cluster with varying degrees of membership. The corresponding objective function is shown in Equation \ref{eq:fuzzy-k-means}, where $p \in (1, +\infty)$ is a hyperparameter that controls how fuzzy the cluster assignments tend to; as $p \rightarrow 1$, the cluster assignments will tend to be less distributed, while as $p \rightarrow +\infty$ the cluster assignments will tend to a uniform distribution. This algorithm is used by 2 (4\%) reviewed works \cite{Sharuee2018, Lazhar2019}.
\begin{equation}
    \label{eq:fuzzy-k-means}
    J(C; D, k, p) =
    \sum_{d \in D}^{}
    \sum_{j=1}^{k}
    \left(
    \left[
    \sum_{z=1}^{k}
    \left(
    \frac{
        \diffnorm{d}{\mu_k}_{2}^{2}
    }{
        \diffnorm{d}{\mu_z}_{2}^{2}
    }
    \right)^\frac{1}{p-1}
    \right]^{-p}
    \cdot
    \hspace{0.15cm}
    \diffnorm{d}{\mu_k}_{2}^{2}
    \right)
\end{equation} 
Employing fuzzy clustering strategies have the benefit of introducing the notion of confidence to the results, since documents next to cluster boundaries will have fuzzier assignments (degrees of membership more distributed) than documents next to the cluster center of mass (with degrees of membership more concentrated on a single cluster).

To cluster documents, the cosine similarity (the cosine of the smallest angle between two embedded documents, see Equation \ref{eq:cosine}) is frequently used rather than the squared Euclidean distance. To achieve this, the classical K-means needs two modifications, which are normalizing the document embeddings $d$ and each centroid estimation $\mu_k$ by the $\ell_{2}$-norm, $d' = d / \Vert d \Vert_{2}$ and $\mu' = \mu / \Vert \mu \Vert_{2}$. Such an approach is known as Spherical K-means \cite{Dhillon2004} and its objective function maximizes the cosine similarity, as shown in Equation \ref{eq:spherical-k-means}. It was used by 2 (4\%) reviewed works \cite{Febrissy2020, Hassani2021}.
\begin{align}
    \label{eq:spherical-k-means}
    J(C; D', k)
    &=
    \sum_{d' \in D'}^{}
    \min\limits_{\mu' \in C}
    \diffnorm{d'}{\mu'}_{2}^{2}
    =
    \sum_{d' \in D'}^{}
    \min\limits_{\mu' \in C}
    (d' - \mu')^{T}(d' - \mu') \notag \\
    & =
    \sum_{d' \in D'}^{}
    \min\limits_{\mu' \in C}
    \left(
    \Vert d' \Vert_{2}^{2} + \Vert \mu' \Vert_{2}^{2} - 2 d'{}^{T}\mu'
    \right)
    =
    \sum_{d' \in D'}^{}
    \min\limits_{\mu' \in C}
    \left(
    2 - 2 \cos(d', \mu')
    \right) \\
    &=
    \sum_{d' \in D'}^{}
    \max\limits_{\mu' \in C}
    \cos(d', \mu') \notag
\end{align}
One major drawback of K-means and its direct variants is that they are too sensitive to ``outliers''. Since it minimizes squared distances to centroids, too distant instances greatly impact the optimal solution. The K-medoids problem also consists of data partitioning to minimize dissimilarities (not necessarily distance functions) from documents to their corresponding \textit{medoids} $m$, which are necessarily existing documents in the corpus ($m \in D$). Equation \ref{eq:k-medoids} formalizes, mathematically, the objective function for this problem as the total negative similarity (or, analogously, the total dissimilarity) for an arbitrary similarity function which, if chosen appropriately, poses an objective function less sensitive to outliers than the traditional K-means objective (which would not be the case if the similarity chosen is the negative squared Euclidean distance, see \cite{Castro2002}).
\begin{equation}
    \label{eq:k-medoids}
    J(M; D) =
    -\sum_{d \in D}^{}
    \max\limits_{m \in M}
    \similarity(d, m)
\end{equation}
Similar to K-means, finding the optimal K-medoids solution is also an NP-hard problem. Hence, only local minima found by heuristics can be efficiently computed, the most popular being the Partition Around Medoids (PAM) algorithm, used by 4 (7\%) reviewed works. Albeit robust against outliers with an appropriate similarity function, PAM is not scalable to large corpora since its time complexity is of $\mathcal{O}(Nk(\lvert D \lvert - k)^{2}) \subseteq \mathcal{O}(Nk\lvert D\lvert^2 + Nk^3)$, a quadratic growth with respect to corpus size $\lvert D\lvert$.

Clustering LARge Applications (CLARA) \cite{Kaufman1986, John1990} is an algorithm that runs PAM on document subsets from corpus $S \subseteq D$, thus scaling the time complexity quadratically only to the subset size $\lvert S\lvert$ but linear growth on corpora size $\lvert D\lvert$, $\mathcal{O}(Nk(\lvert S\lvert - k)^{2} + Nk(\lvert D\lvert - k))$. The traditional recommendation is to choose $\lvert S\lvert \in \mathcal{O}(k)$ such as $\lvert S\lvert = 40 + 2k$, and hence the time complexity becomes $\mathcal{O}(Nk^3 + Nk\lvert D\lvert)$. Another well known PAM modification is the algorithm Clustering Large Applications based on RANdomized Search (CLARANS) \cite{Ng2002}, which adds more randomization to CLARA in order to reduce execution cost.

These algorithms have been continuously optimized since their invention. Recently, \cite{Schubert2021} proposed FasterPAM, consisting of a faster algorithm equivalent to the original PAM (and consequently proposing FasterCLARA and FasterCLARANS since they rely directly on PAM implementation), reducing time complexity of PAM optimization steps up to a factor of $\mathcal{O}(k)$ to $\mathcal{O}(N\lvert D\lvert(\lvert D\lvert - k)) \subseteq \mathcal{O}(N\lvert D\lvert^{2})$, and hence $\mathcal{O}(Nk^{2})$ for FasterCLARA and FasterCLARANS (by choosing $\lvert S\lvert \in \mathcal{O}(k)$). This same work also proposes to at least double the traditional recommended sample size $\lvert S\lvert = 80 + 4k$.

\subsection{Hierarchical Clustering}\label{subsec:hierarchical-clustering}
From 8 (14\%) works that used hierarchical clustering algorithms in their experiments, 7 employed Hierarchical Agglomerative Clustering (HAC) algorithms varying the linkage (how to define the dissimilarity of two clusters given they have multiple objects) from single, complete, group average and Ward's method. While the linkage is a very important hyperparameter that often drastically changes the resulting cluster structure, all the cited linkages can be viewed by the Lance-Williams (LW) unified HAC mathematical framework \cite{Lance1967} shown in Equation \ref{eq:lance-williams}, where $S_z$ denotes the $z$-th partition, $S_i \cup S_j$ denotes a candidate fusion of partitions $S_i$ and $S_j$, $d : S^2 \rightarrow \mathbb{R}$ denotes some dissimilarity measure, $\alpha : \mathbb{N}^3 \rightarrow \mathbb{R}^4$ is a specific weight vector that defines a linkage method, and $n_z = \lvert S_z\lvert$ for brevity. This recurrent formula is optimized greedily by the HAC implementations.
\begin{equation}
    \label{eq:lance-williams}
    \begin{split}
        d(S_i \cup S_j, S_k) =
        \left[
        d(S_i, S_k),
        d(S_j, S_k),
        d(S_i, S_j),
        \lvert d(S_i, S_k) - d(S_j, S_k)\lvert
        \right]^T
        \cdot
        \hspace{0.1cm}
        \alpha(n_i, n_j, n_k)
    \end{split}
\end{equation}
While single linkage (also known as \textit{minimal} or \textit{nearest neighbor} linkage) is the most flexible approach to model arbitrary cluster shapes, local data variations affect the whole cluster structure, and therefore seemingly distinct clusters may end up connected very early by narrow paths in-between clusters. This last property is the chaining effect, where the single linkage is the linkage method affected by it the most among the cited linkages. Complete linkage (also known as \textit{maximal}, \textit{clique} or \textit{farthest neighbor} linkage) is less affected by local variations, but is biased towards hyperspherical-like cluster shapes and tends to fragment large clusters. The (group) average linkage is a compromise between single and complete linkage (\cite[p.284]{Steinbach2003},  \cite[p.559]{Tan2018}). In the LW framework, $\alpha_\text{single}^T = \left[\frac{1}{2}, \frac{1}{2}, 0, -\frac{1}{2}\right]^T$,  $\alpha_\text{complete}^T = \left[\frac{1}{2}, \frac{1}{2}, 0, \frac{1}{2}\right]^T$ and $\alpha_\text{avg}^T = \left[\frac{n_i}{(n_i + n_j)}, \frac{n_j}{(n_i + n_j)}, 0, 0\right]^T$ \cite{Lance1967}.

The traditional Ward's linkage (also known as the Ward's method) \cite{Ward1963, Murtagh2014} corresponds to a special case of the objective function shown in Equation \ref{eq:general-ward} when $p = q = 2$, where $A,B$ are candidate partitions to fusion \cite{Szekely2005}. By using this objective function, the Ward's linkage is generalized to any value of $p \in (0, 2]$ if $q = 2$, which corresponds to optimizing to particular cases of Minkowski distance rather than the usual squared Euclidean distance. It is also valid for $p = q = 1$, corresponding to minimizing the Manhattan (or $\ell_1$) distance \cite{Strauss2017}. In the LW framework, $\alpha_\text{ward}^T = \left[\frac{(n_i + n_k)}{M}, \frac{(n_j + n_k)}{M}, -\frac{n_k}{M}, 0\right]^T$ where $M = n_i + n_j + n_k$ and for every valid Ward's linkage generalization described previously \cite{Wishart1969, Szekely2005, Strauss2017}.
\begin{equation}
    \label{eq:general-ward}
    \begin{split}
    d(A, B; p, q) = 
    \frac{n_a n_b}{n_a + n_b}
    \left(
        \frac{2}{n_a n_b}
        \sum_{i=1}^{n_a}
        \sum_{j=1}^{n_b}
        \diffnorm{a_i}{b_j}_{q}^{p}-\frac{1}{n_a^2}
        \sum_{i,j=1}^{n_a}
        \diffnorm{a_i}{a_j}_{q}^{p}
        - \frac{1}{n_b^2}
        \sum_{i,j=1}^{n_b}
        \diffnorm{b_i}{b_j}_{q}^{p}
    \right)
    \end{split}
\end{equation}

If naively implemented, the worst-case time complexity for every linkage listed in this section is $\mathcal{O}(\lvert D\lvert^{3})$, but it can all be further reduced to $\mathcal{O}(\lvert D\lvert^{2})$ by using more intelligent methods such as Reciprocal Nearest Neighbors Chain (see \cite{Murtagh1983, Murtagh2014, Mullner2011}) or simple heap-based algorithms with worst-case time complexity of $\mathcal{O}(\lvert D \lvert^2\log \lvert D\lvert)$ \cite{Manning2008, Tan2018}. Since every mentioned linkage function has the same worst-case time complexity, it must be chosen accordingly to the task at hand if quadratic running time in relation to corpus size is still acceptable.

\subsection{Density-Based Clustering}\label{subsec:density-based}
Density-based Spatial Clustering of Applications with Noise (DBSCAN) \cite{Ester1996} builds clusters for regions with a high concentration of data or densely populated. It is dependent on tuning two hyperparameters $\epsilon > 0$, which are a fixed minimum radius to decide whether two points in space are considered connected, and $m \in \mathbb{N}$ that is the minimum number of connections a point $d$ needs to satisfy to be considered a \textit{core point}. Core points \textit{reachable} from each other (there is a connected path consisting of only core points between both) are clustered together, alongside all non-core points directly connected to at least one core point (so-called \textit{border points}). Non-core and non-border points are considered outliers and are not clustered.

Three benefits of DBSCAN are its natural capacity of handling outlier documents, it does not assume the number of clusters that must be built, and it can build clusters of arbitrary shapes and volume. The downsides are its inherent dependency of hyperparameter tuning and, since they both fix \textit{a priori} the rules on how the points must be connected, the algorithm cannot appropriately handle data with varying densities \cite[p.569]{Campello2013, Tan2018}. It has a worst-case time complexity of $\mathcal{O}(\lvert D \lvert^{2})$ but only $\mathcal{O}(\lvert D\lvert\log \lvert D\lvert)$ in average. This was used in 2 (4\%) reviewed work experiments \cite{Thaiprayoon2020, Suligoj2015}.

Hierarchical DBSCAN (HDBSCAN) \cite{Campello2013} is a direct extension of the classical DBSCAN algorithm. As its name suggests, it combines both the density and hierarchical clustering approaches, which means it finds densely populated regions but also connects hierarchically those regions by similarity. Thus, HDBSCAN can be thought of as running DBSCAN to infinitely many $\epsilon$ parameters simultaneously. Hence, HDBSCAN does not suffer from variable density in data and is also more insensitive to its hyperparameter selection than the original DBSCAN. Its time complexity was originally $\mathcal{O}(\lvert D\lvert^{2})$ but was improved to approximately $\mathcal{O}(g(D)\lvert D\lvert\log\lvert D\lvert)$ on average by \cite{McInnes2017}, where $g(D)$ represents non-trivial, data-dependent factors to the power up to 10. Nevertheless, \cite{McInnes2017} shows that HDBSCAN running time is comparable to the original DBSCAN in practice. Despite its potential, only one reviewed work used HDBSCAN \cite{grootendorst2020bertopic}.

\subsection{Connection with Factorization Algorithms}\label{subsec:connection}
The NMF algorithm has natural soft (or fuzzy) clustering interpretation due to the non-negativity constraints \cite{Ding05onthe, Kim2008}. More precisely, after the factorization, the matrix $W \in \mathbb{R}_{+}^{n \times k}$ can be interpreted as the membership scores of each document to every latent topic \cite{Febrissy2020, Liu2015, Casalino2017, Morup2012}. A similar analysis can be made to LSA/LSI and pLSA/pLSI due to their connections with NMF and other clustering algorithms \cite{Ding2008, Singh2008}. LDA already gives a document membership score for each latent topic by design, and thus also has a soft-clustering interpretation.

\section{Acknowledgment}\label{sec:acknowledgment}

This work was supported by Brazil's Chamber of Deputies, CAPES, CNPq (grant number: 309858/2021-6), and Google (Latin America Research Awards - first author).


\end{document}